\pdfoutput=1

\documentclass[11pt]{article}
\usepackage[dvipsnames]{xcolor} 
\usepackage[]{acl}

\usepackage{times}
\usepackage{latexsym}

\usepackage[T1]{fontenc}

\usepackage[utf8]{inputenc}

\usepackage{microtype}

\usepackage{inconsolata}

\usepackage{longtable}
\usepackage{multirow}
\usepackage{graphicx}
\usepackage{hyperref}
\usepackage{url}
\usepackage{enumerate}
\usepackage{booktabs}
\usepackage{enumitem}
\usepackage{bbding}
\usepackage{colortbl}
\usepackage{diagbox}
\usepackage{wrapfig}
\usepackage[skins]{tcolorbox}
\usepackage{hyperref}
\usepackage{caption}

\definecolor{b1}{HTML}{1280B0}
\definecolor{b2}{HTML}{25537D}
\colorlet{punct}{red!60!black}
\definecolor{background}{HTML}{EEEEEE}
\definecolor{delim}{RGB}{20,105,176}
\colorlet{numb}{magenta!60!black}

\definecolor{F7E0D5}{RGB}{245,240,255}
\definecolor{mygray}{RGB}{221,221,221}
\definecolor{myblue}{RGB}{36,78,127}
\definecolor{mybrown}{RGB}{132,60,12}
\colorlet{mycolor}{White!0!F7E0D5}
\newcommand{\stitle}[1]{\vspace{1ex} \noindent{\bf #1.}}
\usepackage{xspace}
\newcommand{\alignedcot}{\texttt{AlignedCoT}\xspace}

\newcommand{\figBlue}[1]{\textcolor[RGB]{44,116,179}{#1}}
\newcommand{\figGreen}[1]{\textcolor[RGB]{120,142,105}{#1}}

\hypersetup{
    colorlinks=true,
    linkcolor=black,
    filecolor=black,      
    urlcolor=b1,
    citecolor=black,
}

\tcbuselibrary{breakable}
\title{
\alignedcot: Prompting Large Language Models via \\Native-Speaking Demonstrations
}

\author{
    Zhicheng Yang\textsuperscript{1}~~~
    Yinya Huang\textsuperscript{4}~~~
    Jing Xiong\textsuperscript{6}\\
    {\bf Liang Feng\textsuperscript{5}} ~~
    {\bf Xiaodan Liang\textsuperscript{6,7}~~~}
    {\bf Yiwei Wang\textsuperscript{3}~~~} 
    {\bf Jing Tang\textsuperscript{1,2}}\\
    $^1$The Hong Kong University of Science and Technology (Guangzhou) \\
    $^2$The Hong Kong University of Science and Technology ~
    $^3$University of California, Merced \\
    $^4$City University of Hong Kong ~~  $^5$Chongqing University \\
    $^6$Shenzhen Campus of Sun Yat-sen University ~~
    $^7$MBZUAI \\
    \tt yangzhch6@gmail.com, 
    \tt wangyw.evan@gmail.com,
    \tt yinya.huang@hotmail.com, \\
    \tt xiongj69@mail2.sysu.edu.cn,
    \tt xdliang328@gmail.com,
    \tt jingtang@ust.hk
}

\begin{document}
\maketitle
\begin{abstract}
Large Language Models prompting, such as using in-context demonstrations, is a mainstream technique for invoking LLMs to perform high-performance and solid complex reasoning (e.g., mathematical reasoning, commonsense reasoning), and has the potential for further human-machine collaborative scientific findings. 
However, current LLMs are delicate and elusive in prompt words and styles. 
And there is an unseen gap between LLM understanding and human-written prompts.
This paper introduces \alignedcot, an LLM-acquainted prompting technique that includes proficient ``native-speaking'' in in-context learning for the LLMs.
Specifically, it achieves consistent and correct step-wise prompts in zero-shot scenarios by progressively probing, refining, and formatting the LLM chain of thoughts so that free from handcrafted few-shot demonstrations while maintaining the prompt quality.
We conduct experiments on mathematical reasoning and commonsense reasoning. We find that LLMs with \alignedcot perform significantly superior to them with human-crafted demonstrations. 
We further apply \alignedcot for rewriting the GSM8K training set, resulting in a \textit{GSM8K-Align} dataset. We observe its benefits for retrieval augmented generation.
The code and data can be found at \href{https://github.com/yangzhch6/AlignedCoT}{https://github.com/yangzhch6/AlignedCoT}.
\end{abstract}

\section{Introduction}
Rapidly increasing capabilities of large language models (LLMs) lead to remarkable advances in various NLP tasks \cite{devlin-etal-2019-bert, Radford2018ImprovingLU, raffel_text2text_transformer}. 
However, 
as LLMs scale up,
the computational load of fine-tuning LLMs becomes generally unaffordable.
Meanwhile, in-context learning (ICL) methods exhibit competing performance with fine-tuning \cite{LLM_few_shot, CoT, think_step_by_step}. 
ICL saves the high costs of training LLMs and enjoys high interpretability from the produced reasoning steps. 
These advantages make ICL emerge as a new popular paradigm for NLP, where LLMs make predictions based on in-context demonstrations. 

\begin{figure}[t!] 
    \includegraphics[width=0.99\linewidth]{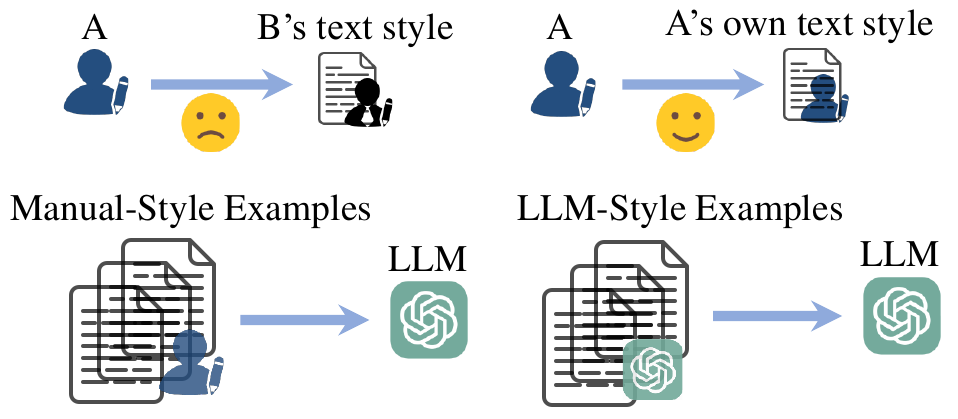}
    \caption{
    A human/machine (A) tends to accept words in her own style (A's own text style) rather than other people’s (B's text style). In this work, we investigate efficient CoT demonstrations by resorting to LLM-learned text habits (an LLM-style).}
    \label{fig:motivation}
    \vspace{-3mm}
\end{figure}

\begin{figure*}[htbp] 
    \includegraphics[width=0.99\linewidth]{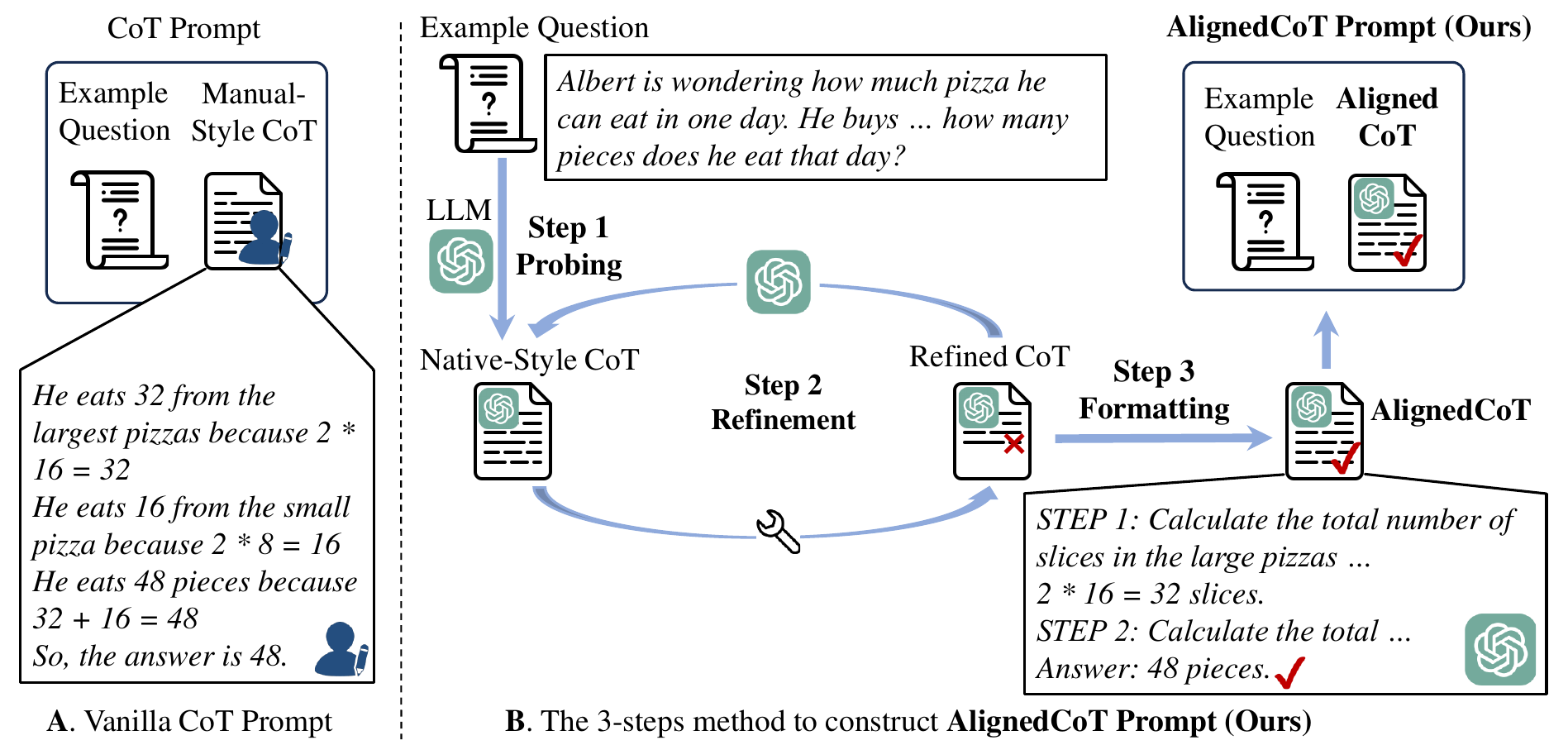}
    \caption{
    \textbf{A}. 
    Existing few-shot demonstrations are conventionally dataset samples or human crafts (``Manual-Style''). As a result, an LLM tends to copy the ``Manual-Style'' format mechanically.
    \textbf{B}. 
    The proposed \alignedcot prompt has zero-shot CoTs with correct and in LLM-acquainted format (``Native-Style’’). The \alignedcot is obtained via three steps: 
    (1) Probing LLM's native style in zero-shot scenarios;
    (2) Refining the generated CoT to correct errors in the first step;
    (3) Formatting the generated CoTs in the first two steps. 
    }
    \label{fig:fig1}
\end{figure*}

A core question in ICL is the selection of effective demonstrations.
\citet{CoT} proposes chain-of-thought (CoT) prompting with
a sequence of short sentences describing intermediate reasoning steps toward final answers.
Originating from CoT,
a line of ICL research \cite{liu-etal-2022-makes, rubin-etal-2022-learning, su2023selective, complex_cot, compositional_icl, li-etal-2023-unified} compose few-shot prompts by selecting examples that are relevant to the input question. 
The other line of work \cite{self_consistency, li-etal-2023-making}
improves LLMs by increasing prompt diversity by sampling reasoning paths multiple times.
Furthermore, another works \cite{yao2023tree, long2023large, hao2023reasoning, zhang2023cumulative}
propose to mimic human cognitive processes.
However, CoT's text style, specifically LLMs' familiarity and proficiency in language use while thinking, and its effect on LLM complex reasoning performances, remain underexplored.

An intuition is that it is more natural for humans to speak in their own style than to imitate others.
As illustrated in Figure \ref{fig:motivation}, a similar situation for LLMs as they may perform better when prompted with their native-style CoT rather than imitating other styles.
However, current CoTs are often human-crafted or come from static dataset samples, resulting in LLMs mimicking the given demonstrations and without their own thinking process. 
Contrastively, LLMs can benefit from a zero-shot manner by generating CoTs
in their own ``native-style'' 
thinking process, which elicits the stored knowledge learned from training (pre-training, SFT, RLHF).


To investigate such ``native-style'' thinking processes, this paper proposes
Aligned Chain-of-Thought (\alignedcot) Prompting.
\alignedcot aims to improve LLM reasoning by aligning the conventional few-shot CoTs to a ``native-style'' zero-shot CoT. 
From the perspective of generalization, the alignment mitigates the disparity between the training and inference; reduces the requirement for extensive model generalization capabilities; and results in performance enhancements.

The proposed \alignedcot operates in steps, as shown
in Figure \ref{fig:fig1}. 
(1) Using each question in the few-shot prompt to query the LLM and generate its native-style CoT in a zero-shot scenario. 
(2) Refining the generated CoT to correct any errors. 
(3) Unifying the CoT text formats, including the format of the final answer and the format of the solution steps. 
We then construct the few-shot prompt using the native-style CoT obtained in the aforementioned three steps to query the LLM.

The contributions of this paper are as follows:
\begin{itemize}
    \item We propose a novel and effective prompting method named \alignedcot, which aligns the CoT text style in few-shot examples to the native style of Large Language Models to improve their reasoning capability. 
    \item We evaluate \alignedcot through extensive experiments, including baseline comparisons and ablation studies. 
    The experimental results show that \alignedcot achieves significant performance improvements and can be easily integrated with other in-context learning methods.
    \item We apply \alignedcot to 
    overwrite the GSM8K dataset
    and provide the \textit{GSM8K-Align} dataset. Empirical results show that \textit{GSM8K-Align} can effectively improve the performance of retrieving augmented methods. 
\end{itemize}

\section{Related Work}

\stitle{Emergent Abilities and Multi-Step Reasoning}
As the amount of computation and data in language models continues to grow, 
advanced capabilities emerge 
\cite{Kaplan2020ScalingLF, wei2022emergent}.
The ability of in-context learning (ICL), that is, to solve the corresponding tasks according to the given few-shot examples, is something that language models are particularly skilled at when scaled up to a certain size \cite{shin-etal-2020-autoprompt, PromptingSurvey}.
Specifically, multi-step reasoning tasks such as math problem solving \cite{cobbe2021training, yang-etal-2022-unbiased, yang-etal-2022-logicsolver, ling-etal-2017-program, wang-etal-2017-deep} and Commonsense Reasoning \cite{suzgun-etal-2023-challenging, srivastava2023beyond, geva-etal-2021-aristotle, talmor-etal-2019-commonsenseqa} witness significant performances growth from larger models, compared to plain tasks such as emotion classification \cite{shin-etal-2020-autoprompt}. 
Moreover, few-shot prompting a language model could outperform itself fine-tuned with a full training set.
In this work, we further investigate the leverage of prompting an LLM for eliciting advanced reasoning capabilities.

\stitle{Chain-of-Thought Reasoning} 
Chain-of-thought prompting \cite{CoT} 
shows that prompting LLMs with intermediate reasoning steps can greatly improve multi-step reasoning ability. Based on this prominent work, further works show that CoT can be improved by various approaches. \citet{self_consistency} propose self-consistency which conducts majority voting by sampling different reasoning paths. Least-to-Most prompting \cite{least_to_most} guides the LLMs to first decompose the original question into small parts and then solve it. Tree-of-Thought \cite{yao2023tree, long2023large} further supports chain-of-thought by solving complex problems in a tree search process. Reasoning via Planning \cite{hao2023reasoning} repositions LLM as both a world model and an inference model, and combines the Monte Carlo Tree Search algorithm to search in a huge inference space. 
It is further observed \cite{think_step_by_step} that 
LLMs are decent zero-shot reasoners and can generate intermediate reasoning steps by simply adding ``Let's think step by step'' before each answer. 
Our work sits in the context of CoT reasoning and proposes a new method to improve reasoning ability in LLMs by aligning CoT text style in few-shot examples to LLM's native style with correct reasoning steps in zero-shot scenarios.

\stitle{Demonstration Design of In-Context Learning} Due to the sensitivity of LLMs to prompts, tasks, and datasets \cite{pmlr-v139-zhao21c, lu-etal-2022-fantastically, su2022selective}, designing prompts and the selection of good examples for in-context learning in few-shot scenarios is a fundamental question \cite{liu-etal-2022-makes}. The vanilla CoT \cite{CoT} prompts LLMs with 8 manually written examples. Based on this, PAL \cite{10.5555/3618408.3618843} converts these examples into programming language statements. 
Complex CoT \cite{complex_cot}
selects examples with the most complex reasoning steps from the training set, which improves multi-step reasoning. 
%
The other line of work
uses retrieval-based methods to extract the most similar and relevant examples in the training set. 
\citet{liu-etal-2022-makes} retrieves semantically similar examples with a test query to formulate its corresponding prompt. 
EPR \cite{rubin-etal-2022-learning} uses an unsupervised retriever to obtain a set of candidate examples. 
CEIL \cite{compositional_icl} leverages contrastive learning to obtain preferred examples.
DQ-Lore \cite{xiong2023dqlore}
leverages Dual Queries and Low-rank approximation Re-ranking for sample selection.
The demonstrations in these works are either from the original training set or handcrafted by humans. 
In this paper, we propose \alignedcot, which leverages LLM to generate correct and unified few-shot demonstrations on their own.

Additionally, Auto-CoT \cite{DBLP:conf/iclr/0001Z0S23} divides the training set into $k$ categories and then selects $k$ samples that are closest to the cluster center, and then 
allows the LLM to automate its own demonstrations.
\citet{nori2023can,zhao2023self} generates self-explanations as in-context exemplars for medical QA.
\citet{wang2023self} built a data-generation pipeline named SP-CoT for open-domain question-answering (OPQA) consisting of 
carefully constructed data generation and composition steps by hand, but
is difficult to transfer to other data fields.
The aforementioned works are limited to specific domains and only discuss generating demonstrations by LLMs themselves.
This paper further delves into error handling in demonstrations and the impact of ``native-style'' CoTs on LLMs performance. 


%
%


\section{\alignedcot Prompting}
Figure~\ref{fig:fig1} shows the proposed \alignedcot.
The main purpose of \alignedcot is to achieve LLM's native-style CoTs without following human-written few-shot demonstrations. Specifically, the native-style CoTs are preferred to be zero-shot CoTs and at the same time have consistent formats and correct steps as stronger substitutions to the few-shot handcrafts.
To achieve this, \alignedcot builds an effective alignment between the two with the following steps:
%
%
Probing LLM's native style in zero-shot scenarios (§\ref{sec:probing}), 
refining the generated CoT to correct errors in the first step (§\ref{sec:refining}), 
and unifying the format of the generated CoTs in the first two steps (§\ref{sec:unify}). 
The obtained \alignedcot is then applied by replacing the CoT demonstrations in the original few-shot prompt to query LLMs. 

\subsection{Probing Native-Style of LLM}
\label{sec:probing}
As shown in Figure \ref{fig:fig1}.A, the chain-of-thought text in the few-shot examples is generally handcrafted \cite{CoT}. 
We consider the text style of handcrafted CoTs as a ``manual-style''. 
Compared with the few-shot scenarios in existing works (which directly use handcrafted examples), LLM does not need to imitate the CoT text that is inconsistent with its own language style in zero-shot scenarios. We refer to the CoT text style generated by LLM in zero-shot scenarios as ``native-style''.
When prompting Large Language Models (LLMs) with ``manual-style'' CoTs, LLMs will follow the formatted demonstrations,
which may not fully exploit the LLM's learning capabilities. 

To bridge this gap, we introduce the first step of our \alignedcot Prompting method, which involves Probing LLM's native-style Chain-of-Thought (CoT) in zero-shot scenarios. 
As illustrated in Figure \ref{fig:fig1}.B, in order to acquire the CoT with native style, we use the magic phrase “Let’s think step by step” proposed by \citet{think_step_by_step} to query LLMs for each example in a given few-shot prompt to generate a CoT that resembles the way it naturally responds to the input question.  

However, the generated native-style CoTs are not always correct. To deal with the errors and inconsistency, we need to proofread the generated content against the ground truth answer in the dataset.

\begin{figure}[t] 
    \includegraphics[width=0.99\linewidth]{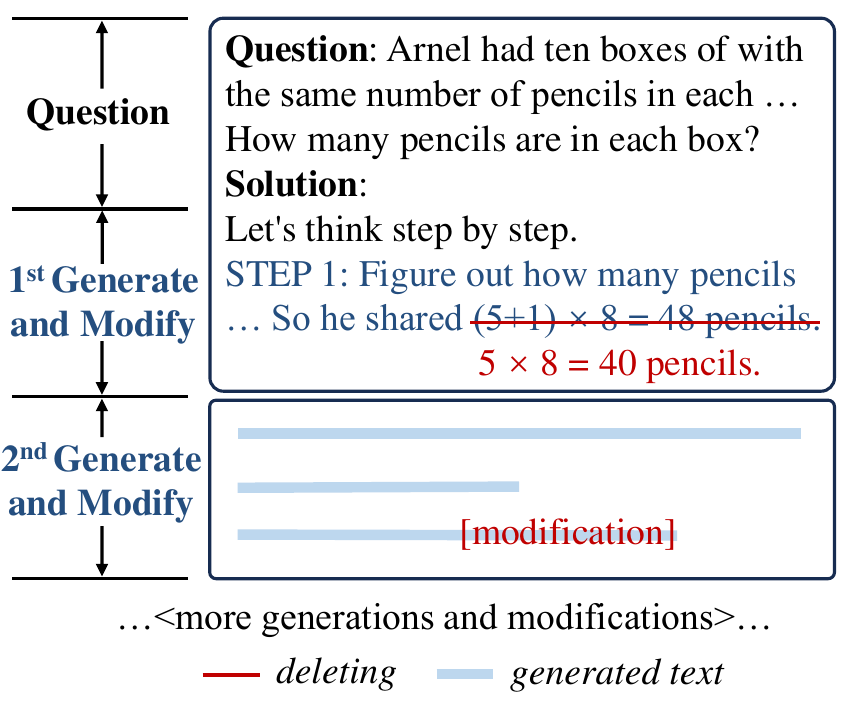}
    \caption{The illustration of our refining process. The modifications in red are annotated manually. We modify the first error each time and then query the LLM to complete the text behind the last modified error.}
    \label{fig:fig2}
\end{figure}

\subsection{Refining CoTs}
\label{sec:refining}
The second critical phase of our \alignedcot Prompting involves the process of refining the CoTs generated in the previous step. This phase is to rectify the inaccuracies 
in the initially generated content. The ultimate goal is to ensure that the CoTs adhere to the highest standards of accuracy, enabling more precise reasoning and response generation by Large Language Models.

The refining process is initiated by identifying and addressing the first encountered error or irrationality in the CoT text.
Subsequently, we harness the capabilities of LLMs to iteratively correct the text, moving forward and completing the answer from the initially modified error position. An example of this process is shown in Figure \ref{fig:fig2}. This iterative approach is executed in the same zero-shot scenario of §\ref{sec:probing}, ensuring that the entire text is rectified while preserving the native style of the LLM's expression. 
Note that
our approach to refinement is designed with a focus on minimalist text modification, which ensures that the generated CoTs are not only error-free but also in harmony with the inherent stylistic nuances of the LLM.

\begin{table*}[t]
\centering
\renewcommand\arraystretch{1.1}
\setlength\tabcolsep{3pt}
\resizebox{1\linewidth}{!}{
\begin{tabular}{c|c|cccccccccc}
\toprule
Model & Prompt & GSM8K & AQUA & SVAMP* & AddSub & SingleEQ & Penguins & Avg \\
\midrule
\multirow{9}*{GPT-3.5-turbo}
~ & CoT w/o \alignedcot & 77.1 & 54.7 & 82.8 & 93.1 & 96.0 & 78.1 & 80.3\\
~ & \cellcolor{mycolor}CoT w/ \textbf{\alignedcot (Ours)} &  
\cellcolor{mycolor}78.7 &
\cellcolor{mycolor}57.1 &
\cellcolor{mycolor}84.8 &
\cellcolor{mycolor}94.9 &
\cellcolor{mycolor}97.6 &
\cellcolor{mycolor}\textbf{87.7} &
\cellcolor{mycolor}83.5 &\\
~ & $\Delta$  & 
\textbf{{\color{White!0!myblue}+1.6$\uparrow$}} & 
\textbf{{\color{White!0!myblue}+2.4$\uparrow$}} & 
\textbf{{\color{White!0!myblue}+2.0$\uparrow$}} & 
\textbf{{\color{White!0!myblue}+1.8$\uparrow$}} & 
\textbf{{\color{White!0!myblue}+1.6$\uparrow$}} & 
\textbf{{\color{White!0!myblue}+9.6$\uparrow$}} & 
\textbf{{\color{White!0!myblue}+3.2$\uparrow$}} \\
\cmidrule{2-9}
~ & Auto-CoT w/o \alignedcot & 78.6 & 50.4 & 81.6 & 92.7 & 96.5 & 80.1 & 80.0 \\
~ & \cellcolor{mycolor}Auto-CoT w/ \textbf{\alignedcot (Ours)} & 
\cellcolor{mycolor}79.8 &
\cellcolor{mycolor}52.0 &
\cellcolor{mycolor}82.3 &
\cellcolor{mycolor}93.9 &
\cellcolor{mycolor}96.5 &
\cellcolor{mycolor}84.9 &
\cellcolor{mycolor}81.6 & \\
~ & $\Delta$  & 
\textbf{{\color{White!0!myblue}+1.2$\uparrow$}} & 
\textbf{{\color{White!0!myblue}+1.6$\uparrow$}} & 
\textbf{{\color{White!0!myblue}+0.7$\uparrow$}} & 
\textbf{{\color{White!0!myblue}+1.2$\uparrow$}} & 
\textbf{{\color{White!0!myblue}+0.0$\uparrow$}} & 
\textbf{{\color{White!0!myblue}+4.8$\uparrow$}} & 
\textbf{{\color{White!0!myblue}+1.6$\uparrow$}} \\

\cmidrule{2-9}
~ & Complex CoT w/o \alignedcot & 79.6 & 55.5 & 82.9 & 93.1 & 96.9 & 81.5 & 81.6\\
~ & \cellcolor{mycolor}Complex CoT w/ \textbf{\alignedcot (Ours)} & 
\cellcolor{mycolor}\textbf{82.4} &
\cellcolor{mycolor}\textbf{57.9} &
\cellcolor{mycolor}\textbf{85.1} &
\cellcolor{mycolor}\textbf{95.2} &
\cellcolor{mycolor}\textbf{98.0} &
\cellcolor{mycolor}86.3 &
\cellcolor{mycolor}\textbf{84.2} & \\
~ & $\Delta$  & 
\textbf{{\color{White!0!myblue}+2.8$\uparrow$}} & 
\textbf{{\color{White!0!myblue}+2.4$\uparrow$}} & 
\textbf{{\color{White!0!myblue}+2.2$\uparrow$}} & 
\textbf{{\color{White!0!myblue}+2.1$\uparrow$}} & 
\textbf{{\color{White!0!myblue}+1.1$\uparrow$}} & 
\textbf{{\color{White!0!myblue}+4.8$\uparrow$}} & 
\textbf{{\color{White!0!myblue}+2.6$\uparrow$}} \\

\bottomrule
\end{tabular}}
\caption{Answer accuracy (\%) of GPT-3.5-Turbo, we compare the performance with or without using our \alignedcot. $\Delta$ indicates performance improvement. 
}
\label{tab:main_result_gpt-3.5-turbo}
\end{table*}

\begin{table*}[t]
\centering
\renewcommand\arraystretch{1.1}
\setlength\tabcolsep{13pt}
\resizebox{1\linewidth}{!}{
\begin{tabular}{c|c|ccccc}
\toprule
Model & Prompt & GSM8K & AQUA & SVAMP* & Penguins & Avg \\
\midrule
\multirow{9}*{GPT-4}
~ & CoT w/o \alignedcot & 93.1 & 72.8 & 94.1 & 96.6 & 89.2 \\
~ & \cellcolor{mycolor}CoT w/ \textbf{\alignedcot (Ours)} & 
\cellcolor{mycolor}94.4 &
\cellcolor{mycolor}\textbf{75.6} &
\cellcolor{mycolor}\textbf{94.8} &
\cellcolor{mycolor}98.6 &
\cellcolor{mycolor}90.9 \\
~ & $\Delta$  & 
\textbf{{\color{White!0!myblue}+1.3$\uparrow$}} & 
\textbf{{\color{White!0!myblue}+2.8$\uparrow$}} & 
\textbf{{\color{White!0!myblue}+0.7$\uparrow$}} & 
\textbf{{\color{White!0!myblue}+2.0$\uparrow$}} & 
\textbf{{\color{White!0!myblue}+1.7$\uparrow$}} \\

\cmidrule{2-7}
~ & Auto-CoT w/o \alignedcot & 93.1 & 72.4 & 93.9 & 97.9 & 89.3 \\
~ & \cellcolor{mycolor}Auto-CoT w/ \textbf{\alignedcot (Ours)} &  
\cellcolor{mycolor} 94.2 &
\cellcolor{mycolor} 73.6 &
\cellcolor{mycolor} 94.4 &
\cellcolor{mycolor} 97.9 &
\cellcolor{mycolor} 90.0 \\
~ & $\Delta$  & 
\textbf{{\color{White!0!myblue}+1.1$\uparrow$}} & 
\textbf{{\color{White!0!myblue}+1.2$\uparrow$}} & 
\textbf{{\color{White!0!myblue}+0.5$\uparrow$}} & 
\textbf{{\color{White!0!myblue}+0.0$\uparrow$}} & 
\textbf{{\color{White!0!myblue}+0.7$\uparrow$}} \\

\cmidrule{2-7}
~ & Complex CoT w/o \alignedcot & 94.4 & 73.6 & 94.2 & 98.6 & 90.2 \\
~ & \cellcolor{mycolor}Complex CoT w/ \textbf{\alignedcot (Ours)} &  
\cellcolor{mycolor}\textbf{95.6} &
\cellcolor{mycolor}74.8 &
\cellcolor{mycolor}94.6 &
\cellcolor{mycolor}\textbf{99.3} &
\cellcolor{mycolor}\textbf{91.1} \\
~ & $\Delta$  & 
\textbf{{\color{White!0!myblue}+1.2$\uparrow$}} & 
\textbf{{\color{White!0!myblue}+1.2$\uparrow$}} & 
\textbf{{\color{White!0!myblue}+0.4$\uparrow$}} & 
\textbf{{\color{White!0!myblue}+0.7$\uparrow$}} & 
\textbf{{\color{White!0!myblue}+0.9$\uparrow$}} \\

\bottomrule
\end{tabular}}
\caption{Answer accuracy (\%) of GPT-4. $\Delta$ indicates performance improvement. 
}
\label{tab:main_result_gpt4}
\end{table*}

\subsection{Unifying the Format of CoTs}
\label{sec:unify}
A crucial aspect that deserves meticulous attention is the consistency of the answer text format and punctuation marks across different examples. This consistency plays a pivotal role in ensuring optimal model performance, as it enables the model to understand and respond to the input queries.

To unify the format of each example, we undertake a manual examination of the generated CoTs from the previous steps. During this evaluation, we focus on two aspects: the format of the answer text and the punctuation marks of solution steps. By meticulously inspecting and revising these elements, we ensure that each CoT conforms to a standardized style, making them more interpretable and consistent in their presentation. 
This approach guarantees that the standardized CoTs maintain a natural and coherent flow, thus enhancing their utility in subsequent reasoning and response generation tasks.



\section{Experiments}
In this section, we first discuss our experimental setting in §\ref{experiment:setting}. In §\ref{experiment:main} and §\ref{experiment:integrate}, we not only show \alignedcot's superior performance in multi-step reasoning but also demonstrate that \alignedcot helps LLM find logical paradoxes better. Furthermore, we conduct ablation study, case study, and more in-depth analysis in §\ref{experiment:ablation} and §\ref{experiment:case}.

\subsection{Experimental Setup}
\label{experiment:setting}
\stitle{Datasets and Language Models} We evaluate our \alignedcot on GSM8K \cite{cobbe2021training}, AQUA \cite{ling-etal-2017-program}, SVAMP\cite{patel-etal-2021-nlp},  AddSub \cite{hosseini-etal-2014-learning}, SingleEQ \cite{koncel-kedziorski-etal-2015-parsing}, and Penguins \cite{suzgun-etal-2023-challenging}. 
%
We choose these datasets because we focus on the multi-step reasoning ability of large language models.
Specifically, there are 7.4k training instances and 1.3k test instances in GSM8K, all samples have manually labeled intermediate problem-solving steps. 
AQUA is a larger mathematical dataset with more difficult samples that are annotated with rationales by human annotators.
AddSub and SingleEQ require the model to have basic computational capabilities.
Penguins contain questions referring to different objects (e.g., Find the name of the oldest penguin).


\stitle{SVAMP and SVAMP*}
SVAMP is a challenging dataset created by applying carefully chosen variations over examples sampled from existing datasets.
However, some questions have logical incorrectness. 
For example: ``\texttt{A waiter had 12 customers. While 15 customers left he got 14 new customers}", 
where the number of customers leaving is actually higher than the number of existing customers, which is impossible to happen. 
Specifically,
we first manually go over and fix the logical paradoxes in the original SVAMP dataset.
There are 7.1\% of the entire dataset that have such logical paradoxes. 
We then have a fixed dataset, denoted as SVAMP*.
We use the fixed SVAMP* for experiments. 
The dataset will be released later.

\stitle{Language Models and Prompts.} 
Our experiments are conducted on GPT-3.5-Turbo \cite{gpt35} and GPT-4 \cite{DBLP:journals/corr/abs-2303-08774} (both use the 0613 version) via the OpenAI API key.  
Our \alignedcot focuses on converting the CoT text style of examples in the given few-shot prompt, it neither requires special sample selection methods nor changes the processes of the prompting methods.
In our experiments, we choose the original CoT prompt proposed by \citet{CoT}, the complex CoT prompt proposed by \citet{complex_cot}, and the prompt extracted by Efficient Prompt Retriever (EPR) \cite{rubin-etal-2022-learning} as the baselines. Following \cite{think_step_by_step}, we add “Let’s think step by step” before the reasoning chains for all baselines to improve the performance.
We show all the prompts we acquired in the Appendix.

\stitle{Compared Methods} 
We compare \alignedcot with previous CoT methods. 
All methods use greedy decoding (i.e. temperature is set to 0). 
The Standard Prompt \cite{CoT} is human-crafted without intermediate reasoning steps, Auto-CoT \cite{DBLP:conf/iclr/0001Z0S23} constructs prompt from a sample pool, and the CoT Prompt \cite{CoT} includes manually designed intermediate steps. The Complex CoT \cite{complex_cot} as a strong baseline consists of the examples with the most complex intermediate reasoning steps. In Complex Prompt, intermediate steps are taken from the source dataset, if the source data does not have intermediate steps, they are manually annotated.

\begin{figure}[tp] 
    \includegraphics[width=0.99\linewidth]{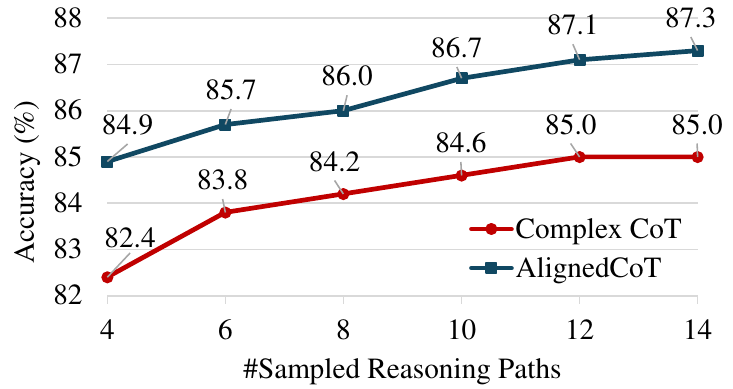}
    \caption{In the case of sampling diverse reasoning paths on GSM8K, our \alignedcot also outperforms Complex CoT.}
    \label{fig:self-consistency}
\end{figure}

\subsection{Main Results}
\label{experiment:main}
Tables \ref{tab:main_result_gpt-3.5-turbo} and \ref{tab:main_result_gpt4} demonstrate the results. 
Note that AddSub and SingleEQ do not serve as benchmarks for GPT-4 since GPT-4 already achieves close to 100\% accuracy on these datasets with CoT prompt.
%
In general, the proposed \alignedcot effectively improves reasoning abilities in LLMs. 
LLMs with CoT empowered by our \alignedcot show an average of +3.2\% and +1.7\% performance improvements for GPT-3.5-turbo and GPT-4, respectively.
Moreover, \alignedcot can work with other CoT methods and have positive synergistic effects. For example, GPT-4 with Auto-CoT empowered by \alignedcot results in an average improvement of 0.9\%. 
GPT-3.5 with Complex CoT empowered by \alignedcot improves on average 2.6\%, and achieves 4.8\% gain on the Penguins dataset. 

We further follow the Self-Consistency method \cite{self_consistency} to compare Complex CoT with our proposed \alignedcot in the case of sampling multiple reasoning paths with temperature set to 0.7, as shown in Figure \ref{fig:self-consistency}. We evaluate the answer accuracy of \texttt{gpt-3.5-turbo} on GSM8K. 
According to the experimental results, the overall performance of our \alignedcot significantly exceeds that of Complex CoT.

\subsection{\alignedcot for Logical Pitfalls}
\label{experiment:data_clean}
\begin{figure*}[t] 
    \includegraphics[width=0.99\linewidth]{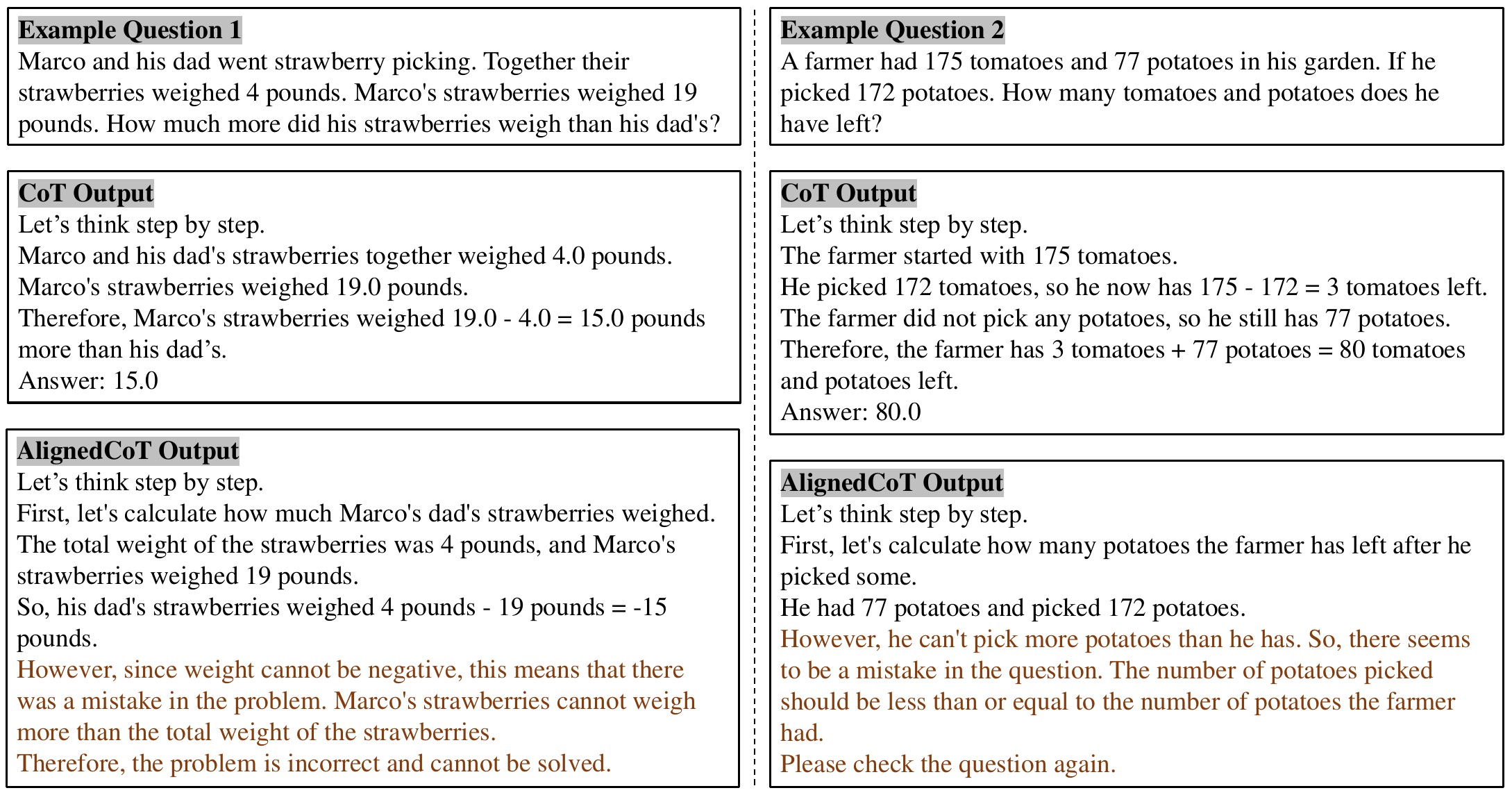}
    \caption{Two cases of logical error detection. The text in {\color{White!0!mybrown}brown} is GPT-4's reasoning process for discovering logical incorrectness.}
    \label{fig:clean_examples}
\end{figure*}

%
To further investigate the capability of \alignedcot for detecting intermediate logical errors, as a qualified counterpart of handcrafted demonstrations,
we conduct experiments on the 7.1\% flaw questions on the original SVAMP, as described in Section~\ref{experiment:setting}. 
The results are shown in Table \ref{tab:logic_error}. 
And some examples are demonstrated in Figure \ref{fig:clean_examples}.
The logical error recognition rate of GPT-3.5-Turbo with \alignedcot is 18.3\%, and the GPT-4 counterpart reaches 78.9\%, significantly outperforming Complex CoT. 
The results suggest that \alignedcot helps LLMs to better spot logical incorrectness in questions. 
%

\begin{table}[t]
\centering
\renewcommand\arraystretch{1.1}
\setlength\tabcolsep{13pt}
\resizebox{1.0\linewidth}{!}{
\begin{tabular}{c|c|c}
\toprule
Model & Complex CoT & \alignedcot \\
\midrule
GPT-3.5-Turbo & 15.5 & 18.3\\
GPT-4 & 28.2 & \textbf{78.9}\\
\bottomrule
\end{tabular}}
\caption{Logical error recognition rate of GPT-3.5-Turbo and GPT-4.
Results are calculated on 7.1\% logically incorrect questions as introduced in Section~\ref{experiment:setting}.
}
\label{tab:logic_error}
\end{table}

\begin{table}[t]
\centering
\resizebox{1\linewidth}{!}{
\begin{tabular}{c|c|c|c}
\toprule
Probing & Refining & Formatting & Answer Accuracy \\ 
\midrule
$\times$ & $\times$ & $\times$ & 79.6 \\
$\surd$ & $\times$ & $\times$ & 80.5 (+0.7)\\
$\surd$ & $\surd$ & $\times$ & 81.5 (+1.7)\\
$\surd$ & $\times$ & $\surd$ & 80.9 (+1.1)\\
$\surd$ & $\surd$ & $\surd$ & \textbf{82.4 ({\color{White!0!myblue}+2.8})} \\
\bottomrule
\end{tabular}}
\caption{Ablation study of different steps in \alignedcot.
Backbone: GPT-3.5-Turbo \cite{gpt35}. Dataset: GSM8K \cite{cobbe2021training}. 
}
\label{tab:ablation}
\end{table}

\begin{table}[t]
\centering
\renewcommand\arraystretch{1.1}
\setlength\tabcolsep{3pt}
\resizebox{0.95\linewidth}{!}{
\begin{tabular}{c|c|c}
\toprule
Method & Prompt & GSM8K \\
\midrule
\multirow{2}*{CoT} & w/o \alignedcot & 28.1 \\
& \cellcolor{mycolor}w/ \textbf{\alignedcot (Ours)} & \cellcolor{mycolor}29.0 ({\color{White!0!myblue}+0.9})\\
\midrule
\multirow{2}*{Complex CoT} & w/o \alignedcot & 28.7 \\
& \cellcolor{mycolor}w/ \textbf{\alignedcot (Ours)}& \cellcolor{mycolor}\textbf{29.8 ({\color{White!0!myblue}+1.1})} \\
\bottomrule
\end{tabular}}
\caption{
Llama2-7b-chat with \alignedcot results.
}
\label{tab:small_model}
\end{table}

\subsection{Ablation Study}
\label{experiment:ablation}
Table \ref{tab:ablation} shows a closer analysis 
of the impact of each step in \alignedcot: Probing, Refining, and Formatting. 
The deletion of all probing, refining, and formatting results in the baseline Complex CoT \cite{complex_cot}, whereas the inclusion of all three is the full \alignedcot.
Experiments use GPT-3.5-Turbo \cite{gpt35} as the backbone and are on GSM8K \cite{cobbe2021training}. 
The results show that solely the first step (Probing) leads to performance improvement of +0.7\%.
Adding the Refining or Formatting step on top of the Probing step, the performance further improves.
When all three steps are deployed simultaneously, our \alignedcot reaches 82.4\%. 
Therefore, the Probing, Refining, and Formatting steps all contribute to \alignedcot for 
multi-step reasoning, and they work best when deployed in combination.
We include further ablation studies on a setting with CORRECT ``native-style'' prompts before the Refining step to find the pure impact of such ``native-style'' prompts without intervention of refinement. Discussions are found in appendix \ref{appendix:native}.

\begin{table}[t]
\centering
\renewcommand\arraystretch{1.1}
\setlength\tabcolsep{3pt}
\resizebox{1\linewidth}{!}{
\begin{tabular}{c|c|c}
\toprule
Retriever & Example Pool & GSM8K \\
\midrule
\multirow{2}*{Random Selection} & Original Data & 76.5 \\
& \cellcolor{mycolor}\textit{GSM8K-Align} & \cellcolor{mycolor}78.0 ({\color{White!0!myblue}+1.5})\\
\midrule
\multirow{3}*{EPR \cite{rubin-etal-2022-learning}} & Original Data & 77.3 \\
& \cellcolor{mygray}\textit{GSM8K-Conv} & \cellcolor{mygray}80.1 ({\color{White!0!myblue}+2.8}) \\
& \cellcolor{mycolor}\textit{GSM8K-Align} & \cellcolor{mycolor}80.9 ({\color{White!0!myblue}\textbf{+3.6}}) \\
\midrule
\multirow{2}*{Complex CoT \cite{complex_cot}} & Original 8-shot & 79.6 \\
& \cellcolor{mycolor}\alignedcot 8-shot& \cellcolor{mycolor}\textbf{82.4 ({\color{White!0!myblue}+2.8})} \\
\bottomrule
\end{tabular}}
\caption{Experimental results of retrieve-based methods on GSM8K. We compare the performance between the Original Data and two types of Aligned Data derived with \alignedcot: \textit{GSM8K-Align} and \textit{GSM8K-Conv}.}
\label{tab:retrieve-results}
\end{table}

\begin{figure*}[t] 
    \includegraphics[width=0.99\linewidth]{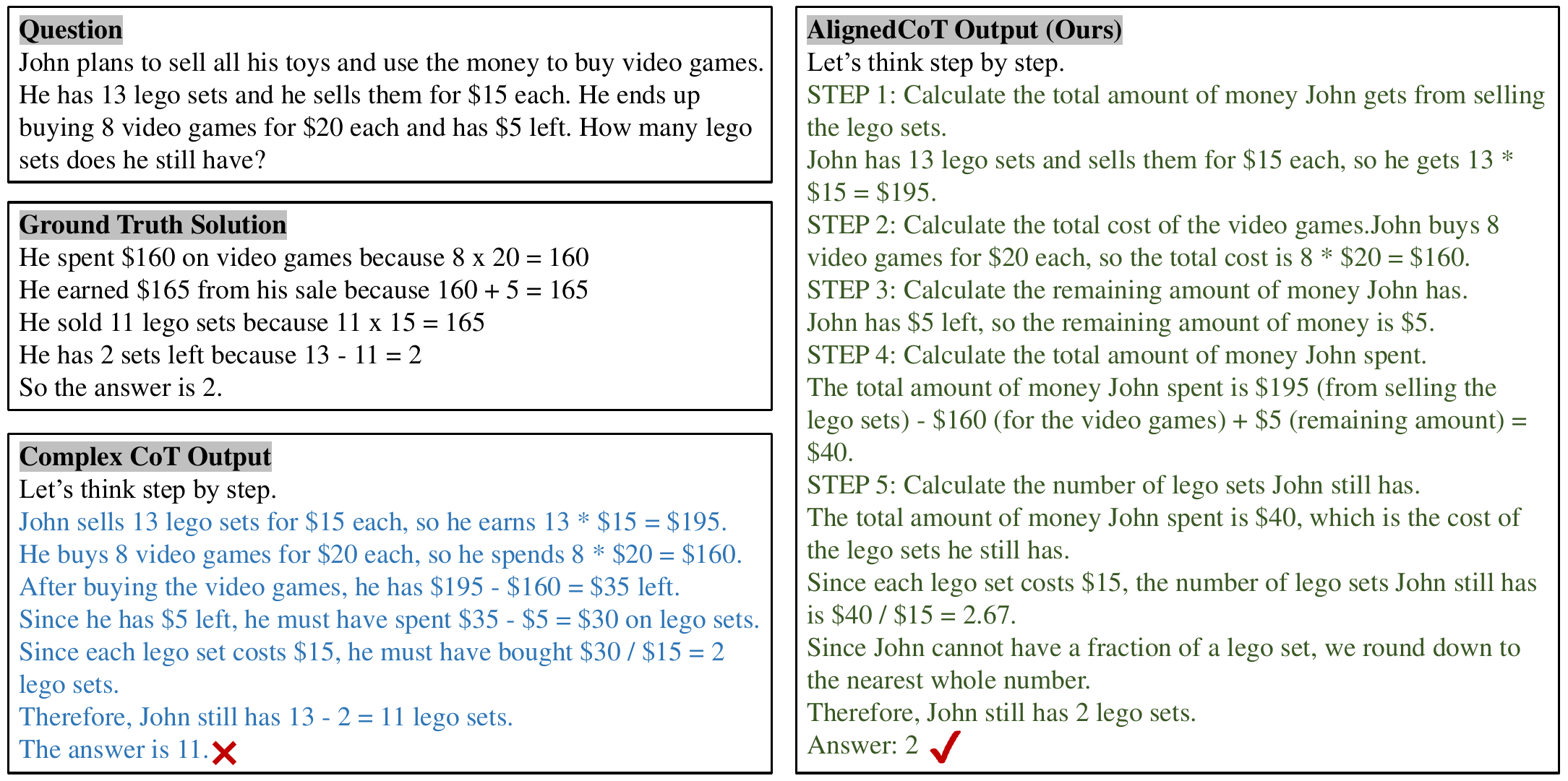}
    \caption{A case of the GSM8K dataset. We show one example in the few-show prompt and one test example from the test split. The ``manual-style'' CoT is colored in \figBlue{blue}, while the ``native-style'' CoT is colored in \figGreen{green}.}
    \label{fig:case}
\end{figure*}

\subsection{\alignedcot with Smaller LMs}
To investigate the generalizability and robustness of our approach,
we further apply \alignedcot to smaller language models. 
Specifically, we conduct experiments with the representative Llama-2-7b-chat \cite{touvron2023llama}.
The results shown in Table \ref{tab:small_model} demonstrate the effectiveness of our \alignedcot across different LLMs and Scales.

\subsection{\alignedcot for RAG}


\label{experiment:integrate}
We further explore the effectiveness of \alignedcot for retrieval-augmented generation (RAG) with in-context learning.
We use \alignedcot to overwrite training data from the original GSM8K training set: 
%
(1) We first use few-shot demonstrations as in Figure~\ref{fig:fig1}.A
to query the LLM to output the native-style CoT for each sample in the training set. 
(2) For those samples that can not be answered correctly in the previous setting, we create a 4-shot prompt with a question, an answer in the original dataset (handcrafted), and a native-style answer from \alignedcot. (\textit{style conversion prompt}.)
(3) For a very small proportion of examples (around 1\%) that can not be answered correctly (1) and (2),
we remove them from the example pool.
Consequently, we obtain the overwritten data
\textit{GSM8K-Align}. Furthermore, we refer to the data generated based on \textit{style conversion prompt} as \textit{GSM8K-Conv}.

We choose Random Selection, Efficient Prompt Retriever (EPR) \cite{rubin-etal-2022-learning}, and Complex CoT \cite{complex_cot} as baselines.
Random Selection method randomly extracts 8 samples from the training set.
Efficient Prompt Retriever (EPR) \cite{rubin-etal-2022-learning} retrieves the most similar training instances as exemplars for a given test case.
Complex CoT \cite{complex_cot} builds prompt by selecting the examples with the most complex reasoning steps in the training set, and therefore also serves as a strong baseline of the retrieve methods. 


Table \ref{tab:retrieve-results} shows the results of \alignedcot for RAG. Further details are in Appendix \ref{appendix:consistent}.
With our Aligned Data, we achieve +1.5\%, +3.6\%, and +2.8\% improvement on Random Selection, EPR, and Complex CoT, respectively. 
Moreover, \textit{GSM8K-Conv} achieves +2.8\% improvement with the EPR retriever, inferior to the \textit{GSM8K-Align} counterpart.
%
The results suggest that \alignedcot is consistently effective for generalized in-context learning settings. The \textit{GSM8K-Align} will be released later. We hope it provides a resource and support to the in-context RAG study. 





\subsection{Case Study}
\label{experiment:case}
Figure \ref{fig:case} demonstrates a comparing case
between LLM's ``native style'' by \alignedcot and ``manual style'' CoT.
Given the test question,
we compare the ground truth answer, LLM ouptut prompted by Complex CoT, and output by \alignedcot CoT, respectively.
Compared to the ``manual-style'' Complex CoT deriving an incorrect answer, the ``native-style'' \alignedcot CoT provides clear intermediate steps for deriving the correct answer. The \alignedcot CoT is also a detailed expansion of the ground truth solution. 
%

\section{Conclusion}
This paper proposes a new zero-shot prompting approach
Aligned Chain-of-Thought (\alignedcot), which elicits LLMs' ``native style'' thinking process for leveraging their profound embedded knowledge for effective multi-step reasoning.  
\alignedcot operates in proving, refining, and formatting to obtain competing ``native-style'' CoTs with handcrafted demonstrations.
Experiments suggest multiple advantages of \alignedcot: 
1) \alignedcot is demonstrated effective for 
multi-step reasoning tasks. Moreover, it can be easily combined with CoT, Complex CoT, and self-consistency and achieve further improvements.
2) \alignedcot enhances LLMs' capability of detecting logical errors in reasoning questions.
3) \alignedcot benefits generalized CoT scenarios including retrieval-augmentated generation.
%


\section{Limitations and Ethical Considerations} 
Due to resource limitations, we are unable to test the proposed approach on more benchmarks. 
Similarly, we can not test the performance based on self-consistency on each benchmark, because the number of samples is proportional to the cost.
%
Moreover, future research endeavors can focus on probing more accurate native styles of LLM and combining the retrieve-based prompting method with native style data.
Future studies can also try to combine \alignedcot with other in-context learning approaches to further improve LLMs.

The data and annotations are collected without personal or confidential information. Therefore, we believe that there is no ethical concern.


\section*{Acknowledgements}
Jing Tang's work is partially supported by National Key R\&D Program of China under Grant No.\ 2023YFF0725100, by the National Natural Science Foundation of China (NSFC) under Grant No.\ 62402410 and U22B2060, by National Language Commission under Grant No.\ WT145‐39, by The Department of Science and Technology of Guangdong Province under Grant No.\ 2023A1515110131, and by Guangzhou Municipal Science and Technology Bureau under Grant No.\ 2023A03J0667 and 2024A04J4454.


\newpage
\onecolumn
\appendix

\section*{Appendix}

\section{Performance Improvement of ``Native-Style'' Prompt}
\label{appendix:native}

In order to further clarify whether the enhancements are mainly attributed to our ``native-style'' CoT or the refinement, we conducted the following experiments:
For the original Complex CoT prompt of GSM8K, we delete the examples that can not be answered correctly by GPT-3.5-Turbo. We denote this prompt as ``ComplexCoT*'' and its Aligned Prompt is ``ComplexCoT*-Align". Further, we also sampled two few-shot prompts, all of the examples can be answered correctly by GPT-3.5-Turbo. They are denoted as ``Prompt-1'' and ``Prompt-2", respectively, and their Aligned Prompt is denoted as ``Align-1'' and ``Align-2".
Under such an experimental setup, we can ensure that our \alignedcot does not involve the refinement step, thus being able to measure the impact of the ``native-style'' prompt on the performance of LLMs.
The results are shown in Table \ref{tab:refine_impact}. 
The experimental results show that without refinement, the ``native-style'' prompt significantly improves model performance. We also believe that the improvement of \alignedcot on model performance mainly comes from the ``native-style'' prompt, and refinement reduces the misleading of wrong examples to LLMs on this basis.

\begin{table}[htbp]
\begin{minipage}[l]{0.34\textwidth}
    \resizebox{1\linewidth}{!}{
    \begin{tabular}{c|c}
    \toprule
    Prompt & GSM8K \\ 
    \midrule
    ComplexCoT* & 79.8 \\
    \cellcolor{mycolor}ComplexCoT*-Align & \cellcolor{mycolor}\textbf{81.9 (+2.3)} \\
    Prompt-1 & 76.8 \\
    \cellcolor{mycolor}Align-1 & \cellcolor{mycolor}77.9 (+1.3) \\
    Prompt-2 & 78.0 \\
    \cellcolor{mycolor}Align-2 & \cellcolor{mycolor}79.5 (+1.5) \\
    \bottomrule
    \end{tabular}}
    \caption{Performance without influence of Refinement.}
    \label{tab:refine_impact}
\end{minipage}\quad\quad\quad\quad
\begin{minipage}[r]{0.55\textwidth}
    \resizebox{1\linewidth}{!}{
    \begin{tabular}{c|cc}
    \toprule 
    Example Pool & Original CoT Acc	& AlignedCoT Acc \\ 
    \midrule
    random selection 1&	76.9 & \cellcolor{mycolor}79.5 (+2.6)\\
    random selection 2&	77.1 & \cellcolor{mycolor}78.4 (+1.3)\\
    random selection 3&	75.5 & \cellcolor{mycolor}76.1 (+0.6)\\
    avg& 76.5& \cellcolor{mycolor}78.0(+1.5) \\
    \bottomrule
    \end{tabular}}
    \caption{Consistent improvements on Random Selection examples.}
    \label{tab:random}
\end{minipage}

\end{table}

\section{Consistent Improvement of \alignedcot across Random Selection Examples}
\label{appendix:consistent}
Our approach has shown consistent performance improvements in the settings of different few-shot examples. In the Random Selection method, we randomly select 8 samples to construct a few-shot prompt. This procedure is repeated 3 times to ensure statistical robustness, following which we computed the average accuracy to accurately gauge the method’s effectiveness. The test accuracy of the 3 prompts obtained from random selection is shown in Table \ref{tab:random}.

\section{Examples of Refinement and Formatting in \alignedcot}
In this section, we show some examples of Refinement (Step 2) and Formatting (Step 3) in our \alignedcot.

\subsection{Example of Refinement}
Figure \ref{fig:fix_app} shows an example of our Refinement process. Human modification part is colored in brown.
\begin{figure*}[htbp] 
    \centering
    \includegraphics[width=1.0\linewidth]{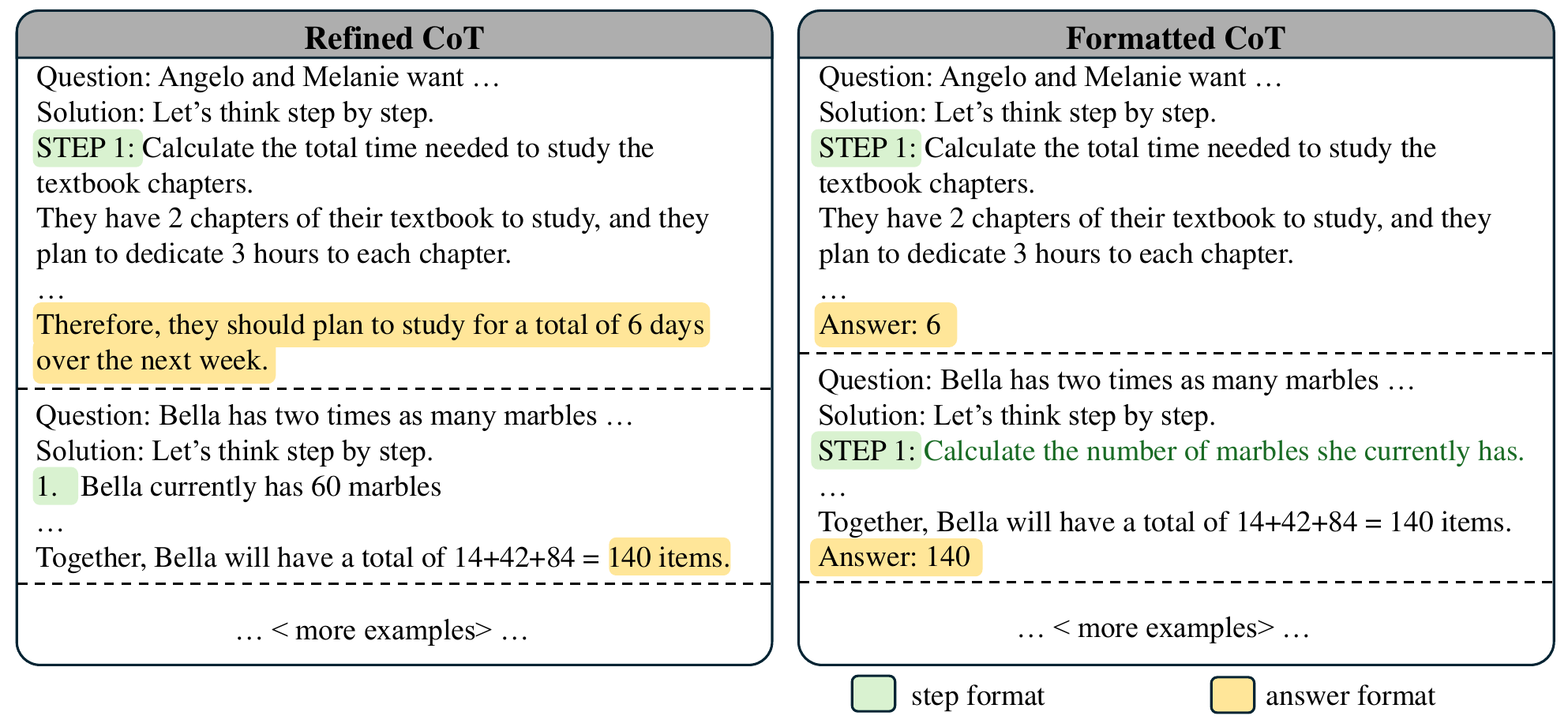}
    \caption{An illustration of formatting answer text and solution steps on GSM8K.}
    \label{fig:format}
\end{figure*}

\begin{figure*}[htbp] 
    \centering
    \includegraphics[width=1\linewidth]{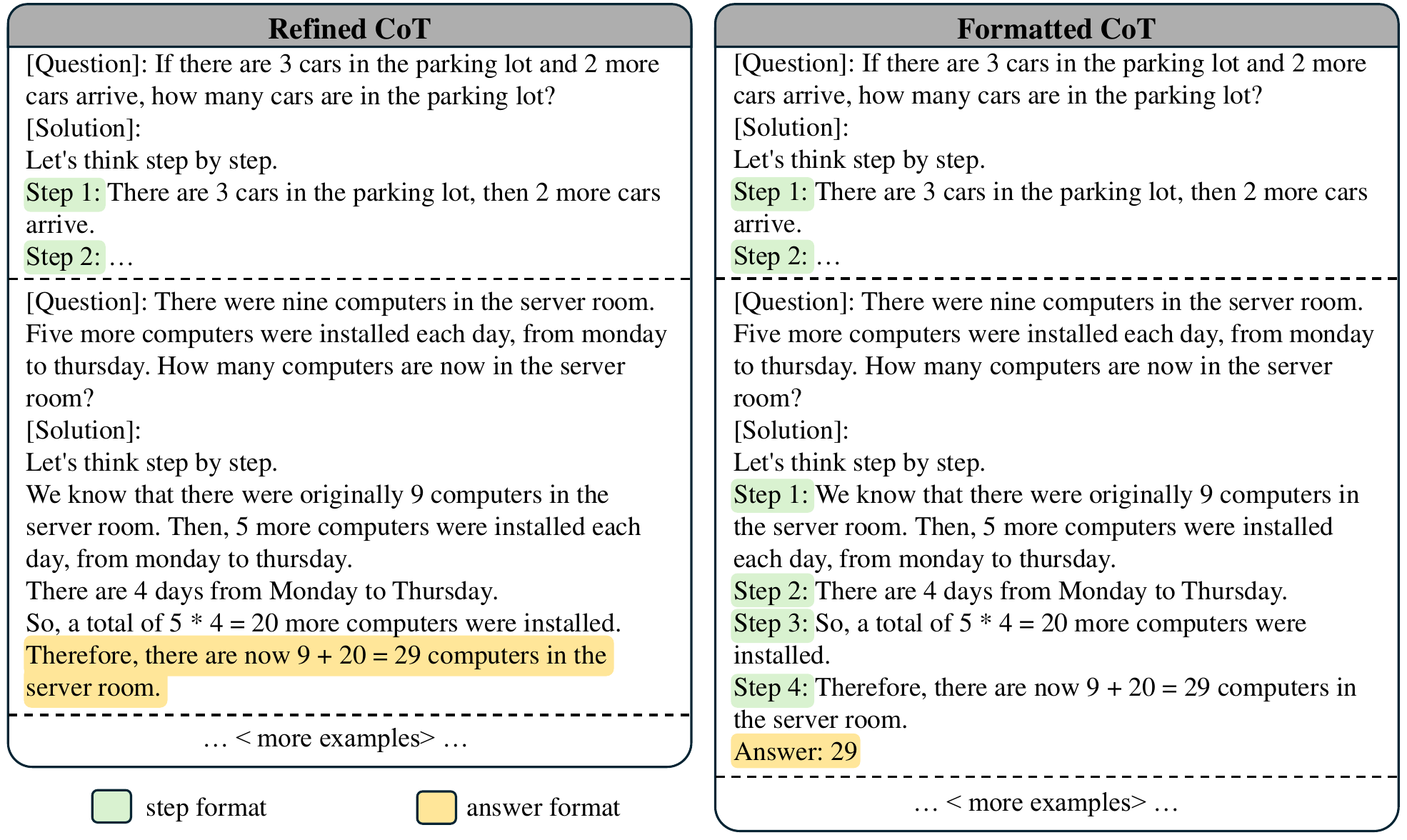}
    \caption{An example of formatting answer text and solution steps on GSM8K.}
    \label{fig:format2}
\end{figure*}

\newpage
\subsection{Example of Formatting}
For Formatting, we aim to unify the format of each examples in the few-shot prompt. This process focus on two aspects: the format of the answer text and the punctuation marks of solution steps. An illustration of this step on GSM8K  is shown in Figure \ref{fig:format}. 'Refined CoT' is the prompt we acquired after Refinement (Step 2). 'Formatted CoT' is the prompt we acquired after Formatting (Step 3), which we use as the final \alignedcot.
More specifically, for the answer text, we unify it as ``Answer:", which also facilitates the extraction of the answer text. For punctuation marks of solution steps, we unify according to the most common and reasonable form in Refined CoT. In Figure \ref{fig:format}, the form is ``STEP N:". If there are no obvious punctuation marks, we don't make changes.
We show some more cases of the Formatting step in Figure \ref{fig:format2}.

\begin{figure*}[htbp] 
    \centering
    \includegraphics[width=0.55\linewidth]{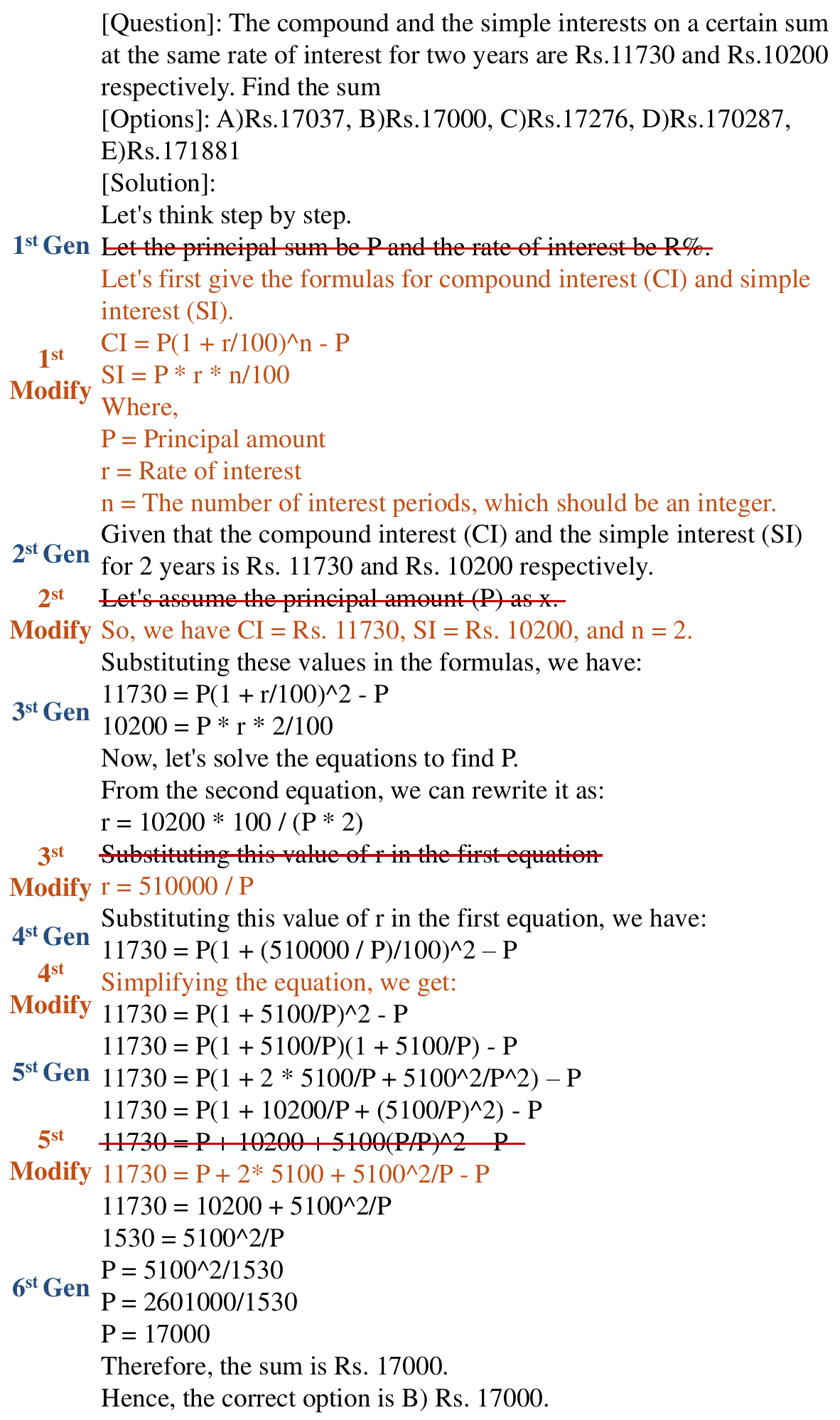}
    \caption{A case of Refinement on Complex CoT for AQUA.}
    \label{fig:fix_app}
\end{figure*}

\section{The Amount of Refinement in \alignedcot}
In our \alignedcot, the proportion of refinement intervention is low, which demonstrates one of the superiorities of our approach.
We present the percentage of samples involved in the Refinement stage for the different few-shot prompts in Table \ref{tab:refine_count_gpt-3.5} and Table \ref{tab:refine_count_gpt-4}. ``Refined/All'' indicates the ratio of the number of refined examples to the total number of examples.
Overall, in our \alignedcot, GPT-3.5-turbo only requires refinement for 11\% of the examples in the prompts of the benchmarks we evaluated, whereas GPT-4 only needs 4\% (two examples).

\begin{table*}[htbp]
\centering
\resizebox{0.85\linewidth}{!}{
\begin{tabular}{c|cccccc|c}
\toprule
Prompt & GSM8K & AQUA & SVAMP* & AddSub & SingleEQ & Penguins & Total \\ 
\midrule
CoT & 0/8 & 1/4 & 0/8 & 0/8 & 0/8 & 0/3 & \multirow{2}*{9/82=11\%}\\
Complex CoT & 2/8 & 3/8 & 1/8 & 1/8 & 1/8 & 0/3\\
\bottomrule
\end{tabular}}
\caption{The proportion of refinement intervention for GPT-3.5-Turbo. The right side of the semicolon '/' represents the number of examples in the prompt, and the left side indicates the number of examples involved in refinement.}
\label{tab:refine_count_gpt-3.5}
\end{table*}

\begin{table*}[htbp]
\centering
\resizebox{0.65\linewidth}{!}{
\begin{tabular}{c|cccc|c}
\toprule
Prompt & GSM8K & AQUA & SVAMP* & Penguins & Total \\ 
\midrule
CoT &         0/8 & 0/4 & 0/8 & 0/3 & \multirow{2}*{2/50=4\%}\\
Complex CoT & 0/8 & 2/8 & 0/8 & 0/3 \\
\bottomrule
\end{tabular}}
\caption{The proportion of refinement intervention for GPT-3.5-Turbo. The right side of the semicolon '/' represents the number of examples in the prompt, and the left side indicates the number of examples involved in refinement.}
\label{tab:refine_count_gpt-4}
\end{table*}

\section{Full Set of Prompts}
We show all the prompts we used in this section. These prompts are also released in our github repo.

\subsection{CoT w/ \alignedcot exemplars for GSM8K, SVAMP, AddSub, and SingleEQ}
\begin{tcolorbox}[breakable, title = {\alignedcot For GPT-3.5-Turbo}]
\small
\textbf{[Question]:} There are 15 trees in the grove. Grove workers will plant trees in the grove today. After they are done, there will be 21 trees. How many trees did the grove workers plant today?\\
\textbf{[Solution]:}\\
Let's think step by step.\\
Step 1: We know that there are originally 15 trees in the grove, and after the grove workers plant trees, there will be 21 trees.\\
Step 2: The workers planted 21 - 15 = 6 trees today.\\
Step 3: Therefore, the grove workers planted 6 trees today.\\
Answer: 6\\
\\
\textbf{[Question]:} If there are 3 cars in the parking lot and 2 more cars arrive, how many cars are in the parking lot?\\
\textbf{[Solution]:}\\
Let's think step by step.\\
Step 1: We know that there are originally 3 cars in the parking lot, and then 2 more cars arrive.\\
Step 2: Therefore, there are 3 + 2 = 5 cars in the parking lot.\\
Answer: 5\\
\\
\textbf{[Question]:} Leah had 32 chocolates and her sister had 42. If they ate 35, how many pieces do they have left in total?\\
\textbf{[Solution]:}\\
Let's think step by step.\\
Step 1: We know that Leah had 32 chocolates and her sister had 42. They ate 35 pieces.\\
Step 2: Leah and her sister had 32 + 42 = 74 chocolates in total.\\
Step 3: After eating 35 chocolates, they have 74 - 35 = 39 chocolates left.\\
Step 4: Therefore, they have 39 chocolates left in total.\\
Answer: 39\\
\\
\textbf{[Question]:} Jason had 20 lollipops. He gave Denny some lollipops. Now Jason has 12 lollipops. How many lollipops did Jason give to Denny?\\
\textbf{[Solution]:}\\
Let's think step by step.\\
Step 1: We know that Jason originally had 20 lollipops, and after giving some to Denny, he now has 12 lollipops.\\
Step 2: Therefore, Jason gave 20 - 12 = 8 lollipops to Denny.\\
Answer: 8\\
\\
\textbf{[Question]:} Shawn has five toys. For Christmas, he got two toys each from his mom and dad. How many toys does he have now?\\
\textbf{[Solution]:}\\
Let's think step by step.\\
Step 1: We know that Shawn originally has 5 toys, and then he gets 2 toys each from his mom and dad for Christmas.\\
Step 2: The total number of toys he got for Christmas is 2 + 2 = 4.\\
Step 3: After Christmas, Shawn has 5 + 4 = 9 toys.\\
Step 4: Therefore, Shawn has 9 toys now.\\
Answer: 9\\
\\
\textbf{[Question]:} There were nine computers in the server room. Five more computers were installed each day, from monday to thursday. How many computers are now in the server room?\\
\textbf{[Solution]:}\\
Let's think step by step.\\
Step 1: We know that there were originally 9 computers in the server room. Then, 5 more computers were installed each day, from monday to thursday.\\
Step 2: There are 4 days from Monday to Thursday.\\
Step 3: So, a total of 5 * 4 = 20 more computers were installed.\\
Step 4: Therefore, there are now 9 + 20 = 29 computers in the server room.\\
Answer: 29\\
\\
\textbf{[Question]:} Michael had 58 golf balls. On tuesday, he lost 23 golf balls. On wednesday, he lost 2 more. How many golf balls did he have at the end of wednesday?\\
\textbf{[Solution]:}\\
Let's think step by step.\\
Step 1: We know that Michael had 58 golf balls. On tuesday, he lost 23 golf balls. On wednesday, he lost 2 more.\\
Step 2: After he lost 23 golf balls, he had 58 - 23 = 35 golf balls left.\\
Step 3: After he lost 2 more, he had 35 - 2 = 33 golf balls left.\\
Step 4: Therefore, at the end of Wednesday, Michael had 33 golf balls.\\
Answer: 33\\
\\
\textbf{[Question]:} Olivia has \$23. She bought five bagels for \$3 each. How much money does she have left?\\
\textbf{[Solution]:}\\
Let's think step by step.\\
Step 1: We know that Olivia has \$23 and she bought five bagels for \$3 each.\\
Step 2: The total cost of the bagels is 5 * \$3 = \$15.\\
Step 3: After buying the bagels, Olivia has \$23 - \$15 = \$8 left.\\
Step 4: Therefore, Olivia has \$8 left.\\
Answer: 8
\end{tcolorbox}

\begin{tcolorbox}[breakable, title = {AlignedCoT For GPT-4}]
\small
\textbf{[Question]}: There are 15 trees in the grove. Grove workers will plant trees in the grove today. After they are done, there will be 21 trees. How many trees did the grove workers plant today?\\
\textbf{[Solution]}: \\
Let's think step by step.\\
First, we know that there were originally 15 trees in the grove.\\
Then, the grove workers planted some trees.\\
After they finished, there were 21 trees in total.\\
So, to find out how many trees they planted, we subtract the original number of trees from the final number of trees.\\
That is, 21 - 15 = 6.\\
So, the grove workers planted 6 trees today.\\
Answer: 6.\\
\\
\textbf{[Question]}: If there are 3 cars in the parking lot and 2 more cars arrive, how many cars are in the parking lot?\\
\textbf{[Solution]}: \\
Let's think step by step.\\
First, there were 3 cars in the parking lot. \\
Then, 2 more cars arrived. \\
So, we add the number of cars that were already there (3) to the number of cars that arrived (2). \\
3 + 2 = 5 \\
So, there are 5 cars in the parking lot.\\
Answer: 5.\\
\\
\textbf{[Question]}: Leah had 32 chocolates and her sister had 42. If they ate 35, how many pieces do they have left in total?\\
\textbf{[Solution]}: \\
Let's think step by step.\\
First, we add the number of chocolates Leah and her sister had together. That is 32 + 42 = 74 chocolates.\\
Then, we subtract the number of chocolates they ate from the total number of chocolates they had. That is 74 - 35 = 39 chocolates.\\
So, they have 39 pieces of chocolates left in total. \\
Answer: 39.\\
\\
\textbf{[Question]}: Jason had 20 lollipops. He gave Denny some lollipops. Now Jason has 12 lollipops. How many lollipops did Jason give to Denny?\\
\textbf{[Solution]}: \\
Let's think step by step.\\
First, Jason had 20 lollipops. Then he gave some to Denny and now he has 12 lollipops. \\
So, to find out how many lollipops Jason gave to Denny, we need to subtract the number of lollipops Jason has now from the number he had at the beginning. \\
So, 20 - 12 = 8. \\
Therefore, Jason gave 8 lollipops to Denny.\\
Answer: 8.\\
\\
\textbf{[Question]}: Shawn has five toys. For Christmas, he got two toys each from his mom and dad. How many toys does he have now?\\
\textbf{[Solution]}: \\
Let's think step by step.\\
First, Shawn had five toys.\\
Then, he got two toys from his mom, so he had 5 + 2 = 7 toys.\\
Finally, he got two more toys from his dad, so he now has 7 + 2 = 9 toys. \\
So, Shawn now has 9 toys.\\
Answer: 9.\\
\\
\textbf{[Question]}: There were nine computers in the server room. Five more computers were installed each day, from monday to thursday. How many computers are now in the server room?\\
\textbf{[Solution]}: \\
Let's think step by step.\\
There were originally 9 computers. \\
Five more computers were installed each day from Monday to Thursday. That's 4 days. So, 5 computers * 4 days = 20 computers were added.\\
So, the total number of computers now in the server room is 9 (original) + 20 (added) = 29 computers.\\
Answer: 29.\\
\\
\textbf{[Question]}: Michael had 58 golf balls. On tuesday, he lost 23 golf balls. On wednesday, he lost 2 more. How many golf balls did he have at the end of wednesday?\\
\textbf{[Solution]}: \\
Let's think step by step.\\
First, Michael lost 23 golf balls on Tuesday. So, 58 - 23 = 35 golf balls remained.\\
Then, he lost 2 more golf balls on Wednesday. So, 35 - 2 = 33 golf balls remained.\\
Therefore, Michael had 33 golf balls at the end of Wednesday.\\
Answer: 33.\\
\\
\textbf{[Question]}: Olivia has \$23. She bought five bagels for \$3 each. How much money does she have left?\\
\textbf{[Solution]}: \\
Let's think step by step.\\
First, we need to find out how much Olivia spent on bagels. She bought 5 bagels for \$3 each, so she spent 5 * \$3 = \$15.\\
Then, we subtract the amount she spent from the amount she had. She had \$23 and spent \$15, so she has \$23 - \$15 = \$8 left.\\
Therefore, Olivia has left \$8.\\
Answer: 8.
\end{tcolorbox}

\subsection{Complex CoT w/ \alignedcot exemplars for GSM8K}
\begin{tcolorbox}[breakable, title = {AlignedCoT For GPT-3.5-Turbo}]
\small
\textbf{[Question]}: Angelo and Melanie want to plan how many hours over the next week they should study together for their test next week. They have 2 chapters of their textbook to study and 4 worksheets to memorize. They figure out that they should dedicate 3 hours to each chapter of their textbook and 1.5 hours for each worksheet. If they plan to study no more than 4 hours each day, how many days should they plan to study total over the next week if they take a 10-minute break every hour, include 3 10-minute snack breaks each day, and 30 minutes for lunch each day?\\
\textbf{[Solution]}:\\
Let's think step by step.\\
STEP 1: Calculate the total time needed to study the textbook chapters.\\
They have 2 chapters of their textbook to study, and they plan to dedicate 3 hours to each chapter.\\
So, the total time needed to study the textbook chapters is 2 * 3 = 6 hours.\\
STEP 2: Calculate the total time needed to memorize the worksheets.\\
They have 4 worksheets to memorize, and they plan to dedicate 1.5 hours for each worksheet.\\
So, the total time needed to memorize the worksheets is 4 * 1.5 = 6 hours.\\
STEP 3: Calculate the total time needed for breaks and meals each day.\\
They take a 10-minute break every hour, so in a 4-hour study session, they will have 4 * 10 = 40 minutes of breaks.\\
They also have 3 10-minute snack breaks each day, so they will have 3 * 10 = 30 minutes of snack breaks.\\
They have 30 minutes for lunch each day.\\
So, the total time needed for breaks and meals each day is 40 + 30 + 30 = 100 minutes, which is equal to 100/60 = 1.67 hours.\\
STEP 4: Calculate the total time needed to study each day.\\
They plan to study no more than 4 hours each day, and they need to subtract the time needed for breaks and meals.\\
So, the total time they can study each day is 4 - 1.67 = 2.33 hours.\\
STEP 5: Calculate the total number of days needed to study.\\
They need a total of 6 hours to study the textbook chapters and 6 hours to memorize the worksheets.\\
So, the total time needed to study is 6 + 6 = 12 hours.\\
Since they can study 2.33 hours each day, they will need 12 / 2.33 = 5.15 days to complete their study.\\
Since they can't study for a fraction of a day, they should plan to study for 6 days.\\
Therefore, they should plan to study for a total of 6 days over the next week.\\
Answer: 6\\
\\
\textbf{[Question]}: Mark's basketball team scores 25 2 pointers, 8 3 pointers and 10 free throws.  Their opponents score double the 2 pointers but half the 3 pointers and free throws.  What's the total number of points scored by both teams added together?\\
\textbf{[Solution]}:\\
Let's think step by step.\\
STEP 1: Calculate the number of points scored by Mark's team.\\
Mark's team scores 25 2 pointers, which means they score 25 * 2 = 50 points from 2 pointers.\\
They also score 8 3 pointers, which means they score 8 * 3 = 24 points from 3 pointers.\\
Finally, they score 10 free throws, which means they score 10 * 1 = 10 points from free throws.\\
Therefore, the total number of points scored by Mark's team is 50 + 24 + 10 = 84 points.\\
STEP 2: Calculate the number of points scored by the opponents.\\
The opponents score double the 2 pointers, which means they score 2 * 50 = 100 points from 2 pointers.\\
They also score half the 3 pointers, which means they score 0.5 * 24 = 12 points from 3 pointers.\\
Finally, they score half the free throws, which means they score 0.5 * 10 = 5 points from free throws.\\
Therefore, the total number of points scored by the opponents is 100 + 12 + 5 = 117 points.\\
STEP 3: Calculate the total number of points scored by both teams.\\
The total number of points scored by both teams is 84 + 117 = 201 points.\\
Answer: 201\\
\\
\textbf{[Question]}: Bella has two times as many marbles as frisbees. She also has 20 more frisbees than deck cards. If she buys 2/5 times more of each item, what would be the total number of the items she will have if she currently has 60 marbles?\\
\textbf{[Solution]}:\\
Let's think step by step.\\
STEP 1: Calculate the number of marbles she currently has.\\
When Bella buys 2/5 times more marbles, she'll have increased the number of marbles by 2/5*60 = 24\\
The total number of marbles she'll have is 60+24 = 84\\
STEP 2: Calculate the number of frisbees she currently has.\\
If Bella currently has 60 marbles, and she has two times as many marbles as frisbees, she has 60/2 = 30 frisbees.\\
If Bella buys 2/5 times more frisbees, she'll have 2/5*30 = 12 more frisbees.\\
The total number of frisbees she'll have will increase to 30+12 = 42\\
STEP 3: Calculate the number of deck cards she currently has.\\
Bella also has 20 more frisbees than deck cards, meaning she has 30-20 = 10 deck cards\\
If she buys 2/5 times more deck cards, she'll have 2/5*10 = 4 more deck cards.\\
The total number of deck cards she'll have is 10+4 = 14\\
STEP 4: Calculate the total number of items she'll have If she buys 2/5 times more of each item.\\
Together, Bella will have a total of 14+42+84 = 140 items\\
Answer: 140\\
\\
\textbf{[Question]}: A group of 4 fruit baskets contains 9 apples, 15 oranges, and 14 bananas in the first three baskets and 2 less of each fruit in the fourth basket. How many fruits are there?\\
\textbf{[Solution]}:\\
Let's think step by step.\\
STEP 1: Figure out the total number of fruits in the first three baskets.\\
In each of the first three basket, there are 9 apples + 15 oranges + 14 bananas = 38 fruits.\\
So, in the first three baskets, there are a total of 38 fruits * 3 = 114 fruits.\\
STEP 2: Figure out the number of fruits in the fourth basket.\\
In the fourth basket, there are 2 less of each fruit compared to the first three baskets.\\
So, there are 9 apples - 2 = 7 apples.\\
There are 15 oranges - 2 = 13 oranges.\\
There are 14 bananas - 2 = 12 bananas.\\
Therefore, in the fourth basket, there are a total of 7 apples + 13 oranges + 12 bananas = 32 fruits.\\
STEP 3: Figure out the total number of fruits in all four baskets.\\
The four baskets together have 114 (total fruits in the first three baskets) + 32 (fruits in the fourth basket) = 146 fruits.\\
Answer: 146\\
\\
\textbf{[Question]}: You can buy 4 apples or 1 watermelon for the same price. You bought 36 fruits evenly split between oranges, apples and watermelons, and the price of 1 orange is \$0.50. How much does 1 apple cost if your total bill was \$66?\\
\textbf{[Solution]}:\\
Let's think step by step.\\
STEP 1: Calculate the number of each kind of fruit you bought.\\
We also know that you bought 36 fruits evenly split between oranges, apples, and watermelons. So, you bought 36/3 = 12 for each kind of fruit.\\
STEP 2: Calculate the total cost of the oranges.\\
We know that the price of 1 orange is \$0.50. So, the total cost of all the oranges is 12 * \$0.50 = \$6.\\
STEP 3: Define the variable.\\
Let's assume the cost of 1 apple is x dollars. \\
STEP 4: Establish an equation according to: the sum of the cost of each kind of fruit (apple, watermelon, and watermelon) equals to the total cost \$66.\\
So, the total cost of all the apples is 12 * x = 12x dollars.\\
We are given that you can buy 4 apples or 1 watermelon for the same price. So, the cost of 1 watermelon is also 4x dollars. \\
The total cost of all the watermelons is 12 * 4x = 48x dollars.\\
We know that the total bill was \$66. \\
Therefore, we can write the equation: \$6 + 12x + 48x = \$66.\\
STEP 5: Solve the equation and get the answer.\\
Simplifying the equation, we get \$6 + 60x = \$66.\\
Subtracting \$6 from both sides, we get 60x = \$60.\\
Dividing both sides by 60, we get x = \$1.\\
So, the cost of 1 apple is \$1.\\
Answer: 1\\
\\
\textbf{[Question]}: Susy goes to a large school with 800 students, while Sarah goes to a smaller school with only 300 students.  At the start of the school year, Susy had 100 social media followers.  She gained 40 new followers in the first week of the school year, half that in the second week, and half of that in the third week.  Sarah only had 50 social media followers at the start of the year, but she gained 90 new followers the first week, a third of that in the second week, and a third of that in the third week.  After three weeks, how many social media followers did the girl with the most total followers have?\\
\textbf{[Solution]}:\\
Let's think step by step.\\
STEP 1: Figure out how many followers Susy has after three weeks.\\
After one week, Susy has 100+40 = 140 followers.\\
In the second week, Susy gains 40/2 = 20 new followers.\\
In the third week, Susy gains 20/2 = 10 new followers.\\
In total, Susy finishes the three weeks with 140+20+10 = 170 total followers.\\
STEP 2: Figure out how many followers Sarah has after three weeks.\\
After one week, Sarah has 50+90 = 140 followers.\\
After the second week, Sarah gains 90/3 = 30 followers.\\
After the third week, Sarah gains 30/3 = 10 followers.\\
So, Sarah finishes the three weeks with 140+30+10 = 180 total followers.\\
STEP 3: Figure out who has the most total followers.\\
Thus, Sarah is the girl with the most total followers with a total of 180.\\
Answer: 180\\
\\
\textbf{[Question]}: Sam bought a dozen boxes, each with 30 highlighter pens inside, for \$10 each box. He rearranged five of these boxes into packages of six highlighters each and sold them for \$3 per package. He sold the rest of the highlighters separately at the rate of three pens for \$2. How much profit did he make in total, in dollars?\\
\textbf{[Solution]}:\\
Let's think step by step.\\
STEP 1: Calculate the total cost of buying the boxes.\\
Sam bought a dozen boxes, which means he bought 12 boxes in total.\\
Each box cost \$10, so the total cost of buying the boxes is 12 * \$10 = \$120.\\
STEP 2: Calculate the total number of highlighters Sam bought.\\
Each box contains 30 highlighter pens, so the total number of highlighters Sam bought is 12 * 30 = 360 highlighters.\\
STEP 3: Calculate the total revenue from selling the packages.\\
Sam rearranged five boxes into packages of six highlighters each, so he made 5 * 30 / 6 = 25 packages.\\
He sold each package for \$3, so the total revenue from selling the packages is 25 * \$3 = \$75.\\
STEP 4: Calculate the total revenue from selling the highlighters separately.\\
Sam sold the rest of the highlighters, which is 360 - (5 * 30) = 210 highlighters, separately.\\
He sold three highlighters for \$2, so the total revenue from selling the highlighters separately is (210/3) * \$2 = \$140.\\
STEP 5: Calculate the total revenue.\\
The total revenue is the sum of the revenue from selling the packages and the revenue from selling the highlighters separately, which is \$75 + \$140 = \$215.\\
STEP 6: Calculate the profit.\\
The profit is the total revenue minus the total cost, which is \$215 - \$120 = \$95.\\
Answer: \$95\\
\\
\textbf{[Question]}: In a certain school, 2/3 of the male students like to play basketball, but only 1/5 of the female students like to play basketball. What percent of the population of the school do not like to play basketball if the ratio of the male to female students is 3:2 and there are 1000 students?\\
\textbf{[Solution]}:\\
Let's think step by step.\\
STEP 1: Figure out the number of male and female students in the school.\\
The students are divided into 3 + 2 = 5 parts where 3 parts are for males and 2 parts are for females.\\
Each part represents 1000/5 = 200 students.\\
So, there are 3 x 200 = 600 males.\\
And there are 2 x 200 = 400 females.\\
STEP 2: Figure out the number of males and females who like to play basketball.\\
2/3 of the male students like to play basketball. Hence, 600 x 2/3 = 400 males play basketball.\\
1/5 of the female students like to play basketball. Hence, 400 x 1/5 = 80 females play basketball.\\
STEP 3: Figure out how many people do not like to play basketball.\\
A total of 400 + 80 = 480 students play basketball.\\
Therefore, 1000 - 480 = 520 do not like to play basketball.\\
STEP 4: Calculate the percentage of the population that do not like to play basketball.\\
The percentage of the school that do not like to play basketball is 520/1000 * 100 = 52\\
Answer: 52
\end{tcolorbox}

\begin{tcolorbox}[breakable, title = {AlignedCoT For GPT-4}]
\small
\textbf{[Question]}: Angelo and Melanie want to plan how many hours over the next week they should study together for their test next week. They have 2 chapters of their textbook to study and 4 worksheets to memorize. They figure out that they should dedicate 3 hours to each chapter of their textbook and 1.5 hours for each worksheet. If they plan to study no more than 4 hours each day, how many days should they plan to study total over the next week if they take a 10-minute break every hour, include 3 10-minute snack breaks each day, and 30 minutes for lunch each day?\\
\textbf{[Solution]}:\\
Let's think step by step.\\
First, calculate the total time they need to study the textbook and worksheets. They need 3 hours for each of the 2 chapters, which is 3*2=6 hours. They also need 1.5 hours for each of the 4 worksheets, which is 1.5*4=6 hours. So, they need a total of 6+6=12 hours to study.\\
Next, calculate the total break time each day. They take a 10-minute break every hour, which is 10*4=40 minutes. They also take 3 10-minute snack breaks, which is 3*10=30 minutes. Plus, they take 30 minutes for lunch. So, they take a total of 40+30+30=100 minutes of breaks each day. Convert this to hours by dividing by 60, which is 100/60=1.67 hours.\\
Subtract the break time from the total hours they can study each day to find out how many hours they can actually study each day. They can study 4 hours each day, but they take 1.67 hours of breaks, so they can actually study 4-1.67=2.33 hours each day.\\
Finally, divide the total hours they need to study by the hours they can actually study each day to find out how many days they need to study. They need to study 12 hours total and can study 2.33 hours each day, so they need to study 12/2.33=5.15 days. Since they can't study for a fraction of a day, they should plan to study for 6 days.\\
Answer: 6\\
\\
\textbf{[Question]}: Mark's basketball team scores 25 2 pointers, 8 3 pointers and 10 free throws.  Their opponents score double the 2 pointers but half the 3 pointers and free throws.  What's the total number of points scored by both teams added together?\\
\textbf{[Solution]}:\\
Let's think step by step.\\
First, calculate the total points scored by Mark's team. A 2 pointer is worth 2 points, a 3 pointer is worth 3 points and a free throw is worth 1 point. So Mark's team scores 25 * 2 + 8 * 3 + 10 * 1 = 50 + 24 + 10 = 84 points.\\
Next, calculate the total points scored by their opponents. They score double the 2 pointers, so that's 25 * 2 * 2 = 100 points. They score half the 3 pointers, so that's 8 * 3 / 2 = 12 points. They also score half the free throws, so that's 10 * 1 / 2 = 5 points. So their opponents score 100 + 12 + 5 = 117 points.\\
Finally, add the points scored by both teams to find the total: 84 + 117 = 201 points.\\
Answer: 201\\
\\
\textbf{[Question]}: Bella has two times as many marbles as frisbees. She also has 20 more frisbees than deck cards. If she buys 2/5 times more of each item, what would be the total number of the items she will have if she currently has 60 marbles?\\
\textbf{[Solution]}:\\
Let's think step by step.\\
First, we know that Bella has two times as many marbles as frisbees. So if she has 60 marbles, she has 60/2 = 30 frisbees.\\
Next, we know that Bella has 20 more frisbees than deck cards. So she has 30 - 20 = 10 deck cards.\\
So currently, Bella has 60 marbles + 30 frisbees + 10 deck cards = 100 items.\\
If Bella buys 2/5 times more of each item, she will have 2/5 * 60 = 24 more marbles, 2/5 * 30 = 12 more frisbees, and 2/5 * 10 = 4 more deck cards.\\
So in total, Bella will have 100 + 24 + 12 + 4 = 140 items.\\
Answer: 140\\
\\
\textbf{[Question]}: A group of 4 fruit baskets contains 9 apples, 15 oranges, and 14 bananas in the first three baskets and 2 less of each fruit in the fourth basket. How many fruits are there?\\
\textbf{[Solution]}:\\
Let's think step by step.\\
First, we need to find out how many fruits are in the first three baskets. In each basket, there are 9 apples, 15 oranges, and 14 bananas. So, in total, there are 9+15+14 = 38 fruits in each basket. Since there are three such baskets, there are 3*38 = 114 fruits in the first three baskets.\\
Next, we need to find out how many fruits are in the fourth basket. We know that there are 2 less of each fruit in the fourth basket. So, there are 9-2 = 7 apples, 15-2 = 13 oranges, and 14-2 = 12 bananas in the fourth basket. In total, there are 7+13+12 = 32 fruits in the fourth basket.\\
Finally, to find out how many fruits are there in total, we need to add the number of fruits in the first three baskets and the number of fruits in the fourth basket. So, there are 114+32 = 146 fruits in total.\\
Answer: 146\\
\\
\textbf{[Question]}: You can buy 4 apples or 1 watermelon for the same price. You bought 36 fruits evenly split between oranges, apples and watermelons, and the price of 1 orange is \$0.50. How much does 1 apple cost if your total bill was \$66?\\
\textbf{[Solution]}:\\
Let's think step by step.\\
First, since the fruits are evenly split, it means you bought 12 of each fruit (12 oranges, 12 apples, and 12 watermelons).\\
Next, we know that the price of 1 orange is \$0.50. So, the total cost for the oranges is 12 * \$0.50 = \$6.\\
We also know that the total bill was \$66. So, the cost for the apples and watermelons is \$66 - \$6 = \$60.\\
Since 4 apples cost the same as 1 watermelon, it means that 12 apples cost the same as 3 watermelons. Therefore, the cost for 12 apples and 12 watermelons (which is 15 fruits in total) is \$60.\\
So, the cost for 1 fruit (either an apple or a watermelon) is \$60 / 15 = \$4.\\
Therefore, since 4 apples cost \$4, 1 apple costs \$4 / 4 = \$1\\
Answer: 1\\
\\
\textbf{[Question]}: Susy goes to a large school with 800 students, while Sarah goes to a smaller school with only 300 students.  At the start of the school year, Susy had 100 social media followers.  She gained 40 new followers in the first week of the school year, half that in the second week, and half of that in the third week.  Sarah only had 50 social media followers at the start of the year, but she gained 90 new followers the first week, a third of that in the second week, and a third of that in the third week.  After three weeks, how many social media followers did the girl with the most total followers have?\\
\textbf{[Solution]}:\\
Let's think step by step.\\
In the first week, Susy gained 40 new followers, so she had 100 + 40 = 140 followers.\\
In the second week, she gained half of 40, which is 40 / 2 = 20 new followers, so she had 140 + 20 = 160 followers.\\
In the third week, she gained half of 20, which is 20 / 2 = 10 new followers, so she had 160 + 10 = 170 followers.\\
On the other hand, Sarah gained 90 new followers in the first week, so she had 50 + 90 = 140 followers.\\
In the second week, she gained a third of 90, which is 90 / 3 = 30 new followers, so she had 140 + 30 = 170 followers.\\
In the third week, she gained a third of 30, which is 30 / 3 = 10 new followers, so she had 170 + 10 = 180 followers.\\
Therefore, the girl with the most total followers after three weeks is Sarah, with 180 followers.\\
Answer: 180\\
\\
\textbf{[Question]}: Sam bought a dozen boxes, each with 30 highlighter pens inside, for \$10 each box. He rearranged five of these boxes into packages of six highlighters each and sold them for \$3 per package. He sold the rest of the highlighters separately at the rate of three pens for \$2. How much profit did he make in total, in dollars?\\
\textbf{[Solution]}:\\
Let's think step by step.\\
First, we calculate the total cost of the boxes. Sam bought a dozen boxes, which is 12 boxes, at \$10 each. So, the total cost is 12 * \$10 = \$20.\\
Next, we calculate how many highlighters Sam has in total. Each box has 30 highlighters, so 12 boxes have 12 * 30 = 360 highlighters.\\
Then, we calculate how many packages of six highlighters Sam made. He rearranged five boxes into packages of six highlighters each. So, he made 5 * 30 / 6 = 25 packages.\\
We calculate how much money Sam made from selling these packages. He sold them for \$3 each, so he made 25 * \$3 = \$75.\\
Next, we calculate how many highlighters Sam sold separately. He sold the rest of the highlighters separately, which is 360 - 5 * 30 = 210 highlighters.\\
We calculate how much money Sam made from selling these highlighters. He sold them at the rate of three pens for \$2, so he made 210 / 3 * \$2 = \$140.\\
Finally, we calculate Sam's profit. His profit is the money he made minus the cost, which is \$75 + \$140 - \$120 = \$95. So, Sam made a profit of \$95\\
Answer: 95\\
\\
\textbf{[Question]}: In a certain school, 2/3 of the male students like to play basketball, but only 1/5 of the female students like to play basketball. What percent of the population of the school do not like to play basketball if the ratio of the male to female students is 3:2 and there are 1000 students?\\
\textbf{[Solution]}:\\
Let's think step by step.\\
First, let's find out how many male and female students there are in the school. The ratio of male to female students is 3:2, which means for every 3 male students, there are 2 female students. \\
So, out of 5 parts (3+2), 3 parts are male and 2 parts are female. \\
Therefore, the number of male students is (3/5)*1000 = 600 and the number of female students is (2/5)*1000 = 400.\\
Next, let's find out how many students like to play basketball. \\
2/3 of the male students like to play basketball, so (2/3)*600 = 400 male students like to play basketball. \\
1/5 of the female students like to play basketball, so (1/5)*400 = 80 female students like to play basketball. \\
So, a total of 400+80 = 480 students like to play basketball.\\
Therefore, the number of students who do not like to play basketball is 1000 - 480 = 520.\\
So, the percentage of students who do not like to play basketball is (520/1000)*100 = 52\%.\\
Answer: 52
\end{tcolorbox}

\subsection{Complex CoT w/ \alignedcot exemplars for SVAMP, AddSub, and SingleEQ}
\begin{tcolorbox}[breakable, title = {Complex CoT we retrieved from SVAMP training set.}]
\small
\textbf{[Question]}: Adam bought 9.0 packages of cat food and 7.0 packages of dog food. Each package of cat food contained 10.0 cans, and each package of dog food contained 5.0 cans. How many more cans of cat food than dog food did Adam buy?\\
\textbf{[Solution]}:\\
Let's think step by step.\\
Adam bought 9.0 packages of cat food, and each package contained 10.0 cans, so he bought a total of 9.0 * 10.0 = 90.0 cans of cat food.\\
Adam also bought 7.0 packages of dog food, and each package contained 5.0 cans, so he bought a total of 7.0 * 5.0 = 35.0 cans of dog food.\\
Therefore, Adam bought 90.0 - 35.0 = 55.0 more cans of cat food than dog food.\\
Answer: 55.0\\
\\
\textbf{[Question]}: In a bag there are 13.0 red marbles, 5.0 blue marbles, and 7.0 green marbles. What percent of the marbles are green?\\
\textbf{[Solution]}:\\
Let's think step by step.\\
To find the percentage of green marbles, we need to divide the number of green marbles by the total number of marbles and then multiply by 100.\\
The total number of marbles is 13.0 + 5.0 + 7.0 = 25.0\\
The number of green marbles is 7.0\\
So, the percentage of green marbles is (7.0/25.0) * 100 = 28\%\\
Answer: 28\\
\\
\textbf{[Question]}: Mika had 20.0 stickers. She bought 26.0 stickers from a store in the mall and got 20.0 stickers for her birthday. Then Mika gave 6.0 of the stickers to her sister and used 58.0 to decorate a greeting card. How many stickers does Mika have left?\\
\textbf{[Solution]}:\\
Let's think step by step.\\
Mika starts with 20.0 stickers.\\
She buys 26.0 stickers, so she now has 20.0 + 26.0 = 46.0 stickers.\\
She also receives 20.0 stickers for her birthday, so she now has 46.0 + 20.0 = 66.0 stickers.\\
Mika gives 6.0 stickers to her sister, so she now has 66.0 - 6.0 = 60.0 stickers.\\
She uses 58.0 stickers to decorate a greeting card, so she now has 60.0 - 58.0 = 2.0 stickers left.\\
Answer: 2.0\\
\\
\textbf{[Question]}: Luke had 20.0 stickers. He bought 12.0 stickers from a store in the mall and got 20.0 stickers for his birthday. Then Luke gave 5.0 of the stickers to his sister and used 8.0 to decorate a greeting card. How many stickers does Luke have left?\\
\textbf{[Solution]}:\\
Let's think step by step.\\
Luke started with 20.0 stickers.\\
He bought 12.0 stickers, so he now has 20.0 + 12.0 = 32.0 stickers.\\
He also received 20.0 stickers for his birthday, so he now has 32.0 + 20.0 = 52.0 stickers.\\
He gave 5.0 stickers to his sister, so he now has 52.0 - 5.0 = 47.0 stickers.\\
He used 8.0 stickers to decorate a greeting card, so he now has 47.0 - 8.0 = 39.0 stickers left.\\
Answer: 39.0\\
\\
\textbf{[Question]}: How many yards of material from a 24.0 yard length of cloth remains after 3.0 pieces each 3.5 yards long and 5.0 pieces each 2.25 long are removed?\\
\textbf{[Solution]}:\\
Let's think step by step.\\
The total length of cloth is 24.0 yards.\\
3 pieces of cloth, each 3.5 yards long, are removed. So, 3 x 3.5 = 10.5 yards of cloth are removed.\\
5 pieces of cloth, each 2.25 yards long, are removed. So, 5 x 2.25 = 11.25 yards of cloth are removed.\\
Therefore, a total of 10.5 + 11.25 = 21.75 yards of cloth are removed.\\
The remaining length of cloth is 24.0 - 21.75 = 2.25 yards.\\
Answer: 2.25\\
\\
\textbf{[Question]}: If Bob can make 100.0 feet of rope in 2.0 hours and Tim can make 75.0 feet of rope in 2.0 hours, how many feet of rope can Bob and Tim working together make in 4.0 hours?\\
\textbf{[Solution]}:\\
Let's think step by step.\\
Bob can make 100.0 feet of rope in 2.0 hours, so his rate of production is 100.0 feet / 2.0 hours = 50.0 feet per hour.\\
Tim can make 75.0 feet of rope in 2.0 hours, so his rate of production is 75.0 feet / 2.0 hours = 37.5 feet per hour.\\
Working together, their combined rate of production is 50.0 feet per hour + 37.5 feet per hour = 87.5 feet per hour.\\
In 4.0 hours, they can make 87.5 feet per hour * 4.0 hours = 350.0 feet of rope.\\
Answer: 350.0\\
\\
\textbf{[Question]}: A neighborhood grocer sells a mix of chocolate and carob candy. The chocolate cost 2.7 dollars a pound and the carob costs 2.55 dollars a pound. If 20.0 pounds of the chocolate and 40.0 pounds of the carob candy are used, what is the cost per pound of the mixture in dollars?\\
\textbf{[Solution]}:\\
Let's think step by step.\\
To find the cost per pound of the mixture, we need to calculate the total cost of the chocolate and carob candy and then divide it by the total weight of the mixture.\\
The total cost of the chocolate candy is 20.0 pounds * \$2.7/pound = \$54.\\
The total cost of the carob candy is 40.0 pounds * \$2.55/pound = \$102.\\
The total weight of the mixture is 20.0 pounds + 40.0 pounds = 60.0 pounds.\\
Therefore, the cost per pound of the mixture is (\$54 + \$102) / 60.0 pounds = \$156 / 60.0 pounds = \$2.60/pound.\\
Answer: 2.60\\
\\
\textbf{[Question]}: Tim got 3.0 nickels and 13.0 dimes for shining shoes, and in his tip jar found 7.0 dimes and 9.0 half - dollars. How much money did Tim get?\\
\textbf{[Solution]}:\\
Let's think step by step.\\
Tim got 3.0 nickels, which is equal to 3 * \$0.05 = \$0.15.\\
Tim also got 13.0 dimes, which is equal to 13 * \$0.10 = \$1.30.\\
In his tip jar, Tim found 7.0 dimes, which is equal to 7 * \$0.10 = \$0.70.\\
Tim also found 9.0 half-dollars, which is equal to 9 * \$0.50 = \$4.50.\\
Therefore, Tim got a total of \$0.15 + \$1.30 + \$0.70 + \$4.50 = \$6.65.\\
Answer: 6.65
\end{tcolorbox}

\begin{tcolorbox}[breakable, title = {AlignedCoT For GPT-3.5-Turbo}]
\small
\textbf{[Question]}: Adam bought 9.0 packages of cat food and 7.0 packages of dog food. Each package of cat food contained 10.0 cans, and each package of dog food contained 5.0 cans. How many more cans of cat food than dog food did Adam buy?\\
\textbf{[Solution]}:\\
Let's think step by step.\\
1. Calculate the total number of cans of cat food. To find the total number of cans of cat food, we need to multiply the number of packages of cat food by the number of cans in each package. 9.0 packages of cat food * 10.0 cans per package = 90.0 cans of cat food.\\
2. Calculate the total number of cans of dog food. To find the total number of cans of dog food, we need to multiply the number of packages of dog food by the number of cans in each package. 7.0 packages of dog food * 5.0 cans per package = 35.0 cans of dog food.\\
3. Calculate the difference in the number of cans of cat food and dog food. To find the difference, we subtract the total number of cans of dog food from the total number of cans of cat food. 90.0 cans of cat food - 35.0 cans of dog food = 55.0 cans.\\
Therefore, Adam bought 55.0 more cans of cat food than dog food.\\
Answer: 55.0\\
\\
\textbf{[Question]}: In a bag there are 13.0 red marbles, 5.0 blue marbles, and 7.0 green marbles. What percent of the marbles are green?\\
\textbf{[Solution]}:\\
Let's think step by step.\\
1. Find the total number of marbles in the bag. Total number of marbles = 13.0 red marbles + 5.0 blue marbles + 7.0 green marbles = 25.0 marbles\\
2. Find the number of green marbles as a fraction of the total number of marbles. Fraction of green marbles = Number of green marbles / Total number of marbles = 7.0 / 25.0\\
3. Convert the fraction to a percentage. Percentage of green marbles = Fraction of green marbles * 100 = (7.0 / 25.0) * 100 = 28.0\%\\
Therefore, 28.0\% of the marbles are green.\\
Answer: 28\\
\\
\textbf{[Question]}: Mika had 20.0 stickers. She bought 26.0 stickers from a store in the mall and got 20.0 stickers for her birthday. Then Mika gave 6.0 of the stickers to her sister and used 58.0 to decorate a greeting card. How many stickers does Mika have left?\\
\textbf{[Solution]}:\\
Let's think step by step.\\
1. Mika had 20.0 stickers.\\
2. She bought 26.0 stickers from a store in the mall, so now she has 20.0 + 26.0 = 46.0 stickers.\\
3. She also got 20.0 stickers for her birthday, so now she has 46.0 + 20.0 = 66.0 stickers.\\
4. Mika gave 6.0 stickers to her sister, so now she has 66.0 - 6.0 = 60.0 stickers.\\
5. Mika used 58.0 stickers to decorate a greeting card, so now she has 60.0 - 58.0 = 2.0 stickers left.\\
Therefore, Mika has 2.0 stickers left.\\
Answer: 2.0\\
\\
\textbf{[Question]}: Luke had 20.0 stickers. He bought 12.0 stickers from a store in the mall and got 20.0 stickers for his birthday. Then Luke gave 5.0 of the stickers to his sister and used 8.0 to decorate a greeting card. How many stickers does Luke have left?\\
\textbf{[Solution]}:\\
Let's think step by step.\\
1. Luke had 20.0 stickers.\\
2. He bought 12.0 stickers from a store in the mall, so he now has 20.0 + 12.0 = 32.0 stickers.\\
3. He also received 20.0 stickers for his birthday, so he now has 32.0 + 20.0 = 52.0 stickers.\\
4. Luke gave 5.0 stickers to his sister, so he now has 52.0 - 5.0 = 47.0 stickers.\\
5. He used 8.0 stickers to decorate a greeting card, so he now has 47.0 - 8.0 = 39.0 stickers left.\\
Therefore, Luke has 39.0 stickers left.\\
Answer: 39.0\\
\\
\textbf{[Question]}: How many yards of material from a 24.0 yard length of cloth remains after 3.0 pieces each 3.5 yards long and 5.0 pieces each 2.25 long are removed?\\
\textbf{[Solution]}:\\
Let's think step by step.\\
1. Start with a 24.0 yard length of cloth.\\
2. Remove 3.0 pieces, each 3.5 yards long. This is a total of 3.0 * 3.5 = 10.5 yards.\\
3. Subtract the length of the removed pieces from the original length: 24.0 - 10.5 = 13.5 yards remaining.\\
4. Remove 5.0 pieces, each 2.25 yards long. This is a total of 5.0 * 2.25 = 11.25 yards.\\
5. Subtract the length of the second set of removed pieces from the remaining length: 13.5 - 11.25 = 2.25 yards remaining.\\
Therefore, after removing the specified pieces, there are 2.25 yards of material remaining from the original 24.0 yard length of cloth.\\
Answer: 2.25\\
\\
\textbf{[Question]}: If Bob can make 100.0 feet of rope in 2.0 hours and Tim can make 75.0 feet of rope in 2.0 hours, how many feet of rope can Bob and Tim working together make in 4.0 hours?\\
\textbf{[Solution]}:\\
Let's think step by step.\\
1. Let's find out how much rope Bob can make in 1 hour. We can do this by dividing the total amount of rope Bob can make in 2 hours (100.0 feet) by the number of hours (2.0 hours).  Bob's rate of making rope = 100.0 feet / 2.0 hours = 50.0 feet/hour\\
2. Let's find out how much rope Tim can make in 1 hour. We can do this by dividing the total amount of rope Tim can make in 2 hours (75.0 feet) by the number of hours (2.0 hours). Tim's rate of making rope = 75.0 feet / 2.0 hours = 37.5 feet/hour\\
3. Let's find out how much rope Bob and Tim can make together in 1 hour. We can do this by adding their individual rates of making rope. Bob and Tim's combined rate of making rope = Bob's rate + Tim's rate = 50.0 feet/hour + 37.5 feet/hour = 87.5 feet/hour\\
4. Let's find out how much rope Bob and Tim can make together in 4 hours. We can do this by multiplying their combined rate of making rope by the number of hours (4.0 hours). Total rope made by Bob and Tim in 4 hours = Combined rate * Number of hours = 87.5 feet/hour * 4.0 hours = 350.0 feet\\
Therefore, Bob and Tim working together can make 350.0 feet of rope in 4.0 hours.\\
Answer: 350.0\\
\\
\textbf{[Question]}: A neighborhood grocer sells a mix of chocolate and carob candy. The chocolate cost 2.7 dollars a pound and the carob costs 2.55 dollars a pound. If 20.0 pounds of the chocolate and 40.0 pounds of the carob candy are used, what is the cost per pound of the mixture in dollars?\\
\textbf{[Solution]}:\\
Let's think step by step.\\
1. We need to find the total cost of the chocolate candy. We can do this by multiplying the cost per pound (\$2.7) by the number of pounds (20.0): Total cost of chocolate candy = \$2.7/pound * 20.0 pounds = \$54.0\\
2. We need to find the total cost of the carob candy. We can do this by multiplying the cost per pound (\$2.55) by the number of pounds (40.0): Total cost of carob candy = \$2.55/pound * 40.0 pounds = \$102.0\\
3. We can find the total cost of the mixture by adding the total cost of the chocolate candy and the total cost of the carob candy: Total cost of mixture = \$54.0 + \$102.0 = \$156.0\\
4. We can find the cost per pound of the mixture by dividing the total cost of the mixture by the total weight of the mixture (20.0 pounds + 40.0 pounds = 60.0 pounds): Cost per pound of mixture = \$156.0 / 60.0 pounds = \$2.6\\
Therefore, the cost per pound of the mixture is \$2.6.\\
Answer: 2.60\\
\\
\textbf{[Question]}: Tim got 3.0 nickels and 13.0 dimes for shining shoes, and in his tip jar found 7.0 dimes and 9.0 half - dollars. How much money did Tim get?\\
\textbf{[Solution]}:\\
Let's think step by step.\\
1. Tim got 3.0 nickels, which means he received 3.0 * \$0.05 = \$0.15 from the nickels.\\
2. Tim also got 13.0 dimes, which means he received 13.0 * \$0.10 = \$1.30 from the dimes.\\
3. In his tip jar, Tim found 7.0 dimes, which means he received an additional 7.0 * \$0.10 = \$0.70 from the dimes in the tip jar.\\
4. Tim also found 9.0 half-dollars, which means he received an additional 9.0 * \$0.50 = \$4.50 from the half-dollars in the tip jar.\\
5. Adding up all the amounts, Tim received \$0.15 + \$1.30 + \$0.70 + \$4.50 = \$6.65 in total.\\
Therefore, Tim get \$6.65.\\
Answer: 6.65
\end{tcolorbox}

\begin{tcolorbox}[breakable, title = {AlignedCoT For GPT-4}]
\small
\textbf{[Question]}: Adam bought 9.0 packages of cat food and 7.0 packages of dog food. Each package of cat food contained 10.0 cans, and each package of dog food contained 5.0 cans. How many more cans of cat food than dog food did Adam buy?\\
\textbf{[Solution]}:\\
Let's think step by step.\\
First, let's calculate how many cans of cat food Adam bought. \\
He bought 9.0 packages of cat food, and each package contained 10.0 cans. \\
So, he bought 9.0 * 10.0 = 90.0 cans of cat food.\\
Next, let's calculate how many cans of dog food Adam bought. \\
He bought 7.0 packages of dog food, and each package contained 5.0 cans. \\
So, he bought 7.0 * 5.0 = 35.0 cans of dog food.\\
Finally, let's calculate how many more cans of cat food than dog food Adam bought. \\
He bought 90.0 cans of cat food and 35.0 cans of dog food. \\
So, he bought 90.0 - 35.0 = 55.0 more cans of cat food than dog food.\\
Answer: 55.0\\
\\
\textbf{[Question]}: In a bag there are 13.0 red marbles, 5.0 blue marbles, and 7.0 green marbles. What percent of the marbles are green?\\
\textbf{[Solution]}:\\
Let's think step by step.\\
First, let's calculate the total number of marbles in the bag. \\
The total number of marbles is 13.0 red marbles + 5.0 blue marbles + 7.0 green marbles = 25.0 marbles.\\
Next, let's calculate the percentage of the marbles that are green. \\
The percentage is the number of green marbles divided by the total number of marbles, multiplied by 100. \\
So, the percentage of green marbles is (7.0 green marbles / 25.0 total marbles) * 100 = 28\%.\\
Answer: 28\\
\\
\textbf{[Question]}: Mika had 20.0 stickers. She bought 26.0 stickers from a store in the mall and got 20.0 stickers for her birthday. Then Mika gave 6.0 of the stickers to her sister and used 58.0 to decorate a greeting card. How many stickers does Mika have left?\\
\textbf{[Solution]}:\\
Let's think step by step.\\
First, let's calculate how many stickers Mika had after she bought some and got some for her birthday. \\
She had 20.0 stickers, bought 26.0 stickers, and got 20.0 stickers for her birthday. \\
So, she had 20.0 + 26.0 + 20.0 = 66.0 stickers.\\
Next, let's calculate how many stickers Mika had left after she gave some to her sister and used some to decorate a greeting card. \\
She gave 6.0 stickers to her sister and used 58.0 stickers. \\
So, she had 66.0 - 6.0 - 58.0 = 2.0 stickers left.\\
Answer: 2.0\\
\\
\textbf{[Question]}: Luke had 20.0 stickers. He bought 12.0 stickers from a store in the mall and got 20.0 stickers for his birthday. Then Luke gave 5.0 of the stickers to his sister and used 8.0 to decorate a greeting card. How many stickers does Luke have left?\\
\textbf{[Solution]}:\\
Let's think step by step.\\
First, let's calculate how many stickers Luke had after he bought some and got some for his birthday. \\
He had 20.0 stickers, bought 12.0 stickers, and got 20.0 stickers for his birthday. \\
So, he had 20.0 + 12.0 + 20.0 = 52.0 stickers.\\
Next, let's calculate how many stickers Luke had left after he gave some to his sister and used some to decorate a greeting card. \\
He gave 5.0 stickers to his sister and used 8.0 stickers. \\
So, he had 52.0 - 5.0 - 8.0 = 39.0 stickers left.\\
Answer: 39.0\\
\\
\textbf{[Question]}: How many yards of material from a 24.0 yard length of cloth remains after 3.0 pieces each 3.5 yards long and 5.0 pieces each 2.25 long are removed?\\
\textbf{[Solution]}:\\
Let's think step by step.\\
First, let's calculate how much material is used for the 3.0 pieces each 3.5 yards long. \\
The total length of these pieces is 3.0 pieces * 3.5 yards/piece = 10.5 yards.\\
Next, let's calculate how much material is used for the 5.0 pieces each 2.25 yards long. \\
The total length of these pieces is 5.0 pieces * 2.25 yards/piece = 11.25 yards.\\
So, the total length of material used is 10.5 yards + 11.25 yards = 21.75 yards.\\
Finally, let's calculate how much material remains from the 24.0 yard length of cloth. \\
The remaining length is 24.0 yards - 21.75 yards = 2.25 yards.\\
Answer: 2.25\\
\\
\textbf{[Question]}: If Bob can make 100.0 feet of rope in 2.0 hours and Tim can make 75.0 feet of rope in 2.0 hours, how many feet of rope can Bob and Tim working together make in 4.0 hours?\\
\textbf{[Solution]}:\\
Let's think step by step.\\
First, let's calculate how much rope Bob and Tim can each make in one hour. \\
Bob can make 100.0 feet / 2.0 hours = 50.0 feet/hour.\\
Tim can make 75.0 feet / 2.0 hours = 37.5 feet/hour.\\
Next, let's calculate how much rope Bob and Tim can make together in one hour. \\
Working together, they can make 50.0 feet/hour + 37.5 feet/hour = 87.5 feet/hour.\\
Finally, let's calculate how much rope Bob and Tim can make together in 4.0 hours. \\
In 4.0 hours, they can make 87.5 feet/hour * 4.0 hours = 350.0 feet.\\
Answer: 350.0\\
\\
\textbf{[Question]}: A neighborhood grocer sells a mix of chocolate and carob candy. The chocolate cost 2.7 dollars a pound and the carob costs 2.55 dollars a pound. If 20.0 pounds of the chocolate and 40.0 pounds of the carob candy are used, what is the cost per pound of the mixture in dollars?\\
\textbf{[Solution]}:\\
Let's think step by step.\\
First, let's calculate the total cost of the chocolate and carob candy. \\
The cost of the chocolate is 20.0 pounds * 2.7 dollars/pound = 54 dollars.\\
The cost of the carob candy is 40.0 pounds * 2.55 dollars/pound = 102 dollars.\\
So, the total cost of the mixture is 54 dollars + 102 dollars = 156 dollars.\\
Next, let's calculate the total weight of the mixture. \\
The total weight is 20.0 pounds + 40.0 pounds = 60.0 pounds.\\
Finally, let's calculate the cost per pound of the mixture. \\
The cost per pound is the total cost divided by the total weight. \\
So, the cost per pound is 156 dollars / 60.0 pounds = 2.6 dollars/pound.\\
Answer: 2.6\\
\\
\textbf{[Question]}: Tim got 3.0 nickels and 13.0 dimes for shining shoes, and in his tip jar found 7.0 dimes and 9.0 half - dollars. How much money did Tim get?\\
\textbf{[Solution]}:\\
Let's think step by step.\\
First, let's calculate how much money Tim got from nickels. \\
A nickel is worth 5 cents. So, 3.0 nickels are worth 3.0 * 5 = 15 cents.\\
Next, let's calculate how much money Tim got from dimes. \\
A dime is worth 10 cents. So, 13.0 dimes from shining shoes are worth 13.0 * 10 = 130 cents.\\
And 7.0 dimes from the tip jar are worth 7.0 * 10 = 70 cents.\\
Then, let's calculate how much money Tim got from half-dollars. \\
A half-dollar is worth 50 cents. So, 9.0 half-dollars are worth 9.0 * 50 = 450 cents.\\
Finally, let's add up all the money Tim got.\\
He got 15 cents from nickels, 130 cents from dimes for shining shoes, 70 cents from dimes in the tip jar, and 450 cents from half-dollars.\\
So, the total amount of money Tim got is 15 + 130 + 70 + 450 = 665 cents.\\
Since 100 cents make a dollar, Tim got 665 / 100 = \$6.65.\\
Answer: 6.65
\end{tcolorbox}

\subsection{CoT w/ \alignedcot exemplars for AQUA}
\begin{tcolorbox}[breakable, title = {AlignedCoT For GPT-3.5-Turbo}]
\small
\textbf{[Question]}: John found that the average of 15 numbers is 40. If 10 is added to each number then the mean of the numbers is?\\
\textbf{[Options]}: A)50, B)45, C)65, D)78, E)64\\
\textbf{[Solution]}: \\
First, Let's find the sum of the 15 numbers.\\
Since the average of the 15 numbers is 40, the sum of the 15 numbers is 15 * 40 = 600.\\
Next, let's add 10 to each number.\\
If we add 10 to each number, the sum of the numbers will increase by 10 * 15 = 150.\\
Then, let's find the new average.\\
The new sum of the numbers is 600 + 150 = 750.\\
Since there are still 15 numbers, the new average is 750 / 15 = 50.\\
Therefore, the mean of the numbers after adding 10 to each number is 50.\\
Answer: A\\
\\
\textbf{[Question]}: If a / b = 3/4 and 8a + 5b = 22,then find the value of a.\\
\textbf{[Options]}: A)1/2, B)3/2, C)5/2, D)4/2, E)7/2\\
\textbf{[Solution]}: \\
Let's think step by step. \\
First, let's solve the equation a/b = 3/4 for a. We can do this by cross-multiplying:\\
4a = 3b\\
Next, let's substitute this value of a into the second equation 8a + 5b = 22:\\
8(3b/4) + 5b = 22\\
Multiplying through by 4 to get rid of the fraction:\\
24b + 20b = 88\\
Combining like terms:\\
44b = 88\\
Dividing both sides by 44:\\
b = 2\\
Now, let's substitute this value of b back into the equation 4a = 3b:\\
4a = 3(2)\\
4a = 6\\
Dividing both sides by 4:\\
a = 6/4\\
Simplifying:\\
a = 3/2\\
Therefore, the value of a is 3/2. \\
Answer: B\\
\\
\textbf{[Question]}: A person is traveling at 20 km/hr and reached his destiny in 2.5 hr then find the distance?\\
\textbf{[Options]}: A)53 km, B)55 km, C)52 km, D)60 km, E)50 km\\
\textbf{[Solution]}: \\
Let's think step by step. \\
Given:\\
Speed = 20 km/hr\\
Time = 2.5 hr\\
To find the distance, we can use the formula:\\
Distance = Speed x Time\\
Substituting the given values:\\
Distance = 20 km/hr x 2.5 hr\\
Calculating:\\
Distance = 50 km\\
Therefore, the distance is 50 km.\\
Answer: E\\
\\
\textbf{[Question]}: How many keystrokes are needed to type the numbers from 1 to 500?\\
\textbf{[Options]}: A)1156, B)1392, C)1480, D)1562, E)1788\\
\textbf{[Solution]}: \\
Let's think step by step.\\
For the numbers from 1 to 9, 1 keystroke is needed for each number. There are 9 numbers from 1 to 9, so 9 keystrokes are needed.\\
For the numbers from 10 to 99, 2 keystrokes are needed for each number. There are 90 numbers from 10 to 99, so 90 * 2 = 180 keystrokes are needed.\\
For the numbers from 100 to 500, 3 keystrokes are needed for each number. There are 401 numbers from 100 to 500, so 401 * 3 = 1203 keystrokes are needed.\\
Adding up all the keystrokes, we get 9 + 180 + 1203 = 1392 keystrokes.\\
Therefore, the 1392 keystrokes are needed to type.\\
Answer: B
\end{tcolorbox}

\begin{tcolorbox}[breakable, title = {AlignedCoT For GPT-4}]
\small
\textbf{[Question]}: John found that the average of 15 numbers is 40. If 10 is added to each number then the mean of the numbers is?\\
\textbf{[Options]}: A)50, B)45, C)65, D)78, E)64\\
\textbf{[Solution]}: \\
Let's think step by step.\\
Step 1: We know the average of 15 numbers is 40, which means the total sum of these numbers is 15 * 40 = 600.\\
Step 2: If 10 is added to each number, the total sum increases by 10 * 15 = 150.\\
Step 3: So, the new total sum is 600 + 150 = 750.\\
Step 4: The new average or mean is the total sum divided by the number of numbers, which is 750/15 = 50.\\
So, the new mean of the numbers is 50.\\
Answer: A\\
\\
\textbf{[Question]}: If a / b = 3/4 and 8a + 5b = 22,then find the value of a.\\
\textbf{[Options]}: A)1/2, B)3/2, C)5/2, D)4/2, E)7/2\\
\textbf{[Solution]}: \\
Let's think step by step.\\
Step 1: From the first equation, we can express a in terms of b. That is a = 3b/4.\\
Step 2: Substitute a = 3b/4 into the second equation. We get 8(3b/4) + 5b = 22, which simplifies to 6b + 5b = 22.\\
Step 3: Combine like terms to get 11b = 22.\\
Step 4: Solve for b by dividing both sides by 11. We get b = 2.\\
Step 5: Substitute b = 2 into the equation a = 3b/4 to get a = 3(2)/4 = 3/2.\\
So, the value of a is 3/2. \\
Answer: B\\
\\
\textbf{[Question]}: A person is traveling at 20 km/hr and reached his destiny in 2.5 hr then find the distance?\\
\textbf{[Options]}: A)53 km, B)55 km, C)52 km, D)60 km, E)50 km\\
\textbf{[Solution]}: \\
Let's think step by step.\\
Step 1: We know the speed of the person which is 20 km/hr.\\
Step 2: We also know the time taken by the person to reach his destination which is 2.5 hr.\\
Step 3: We need to find the distance. We know that distance = speed * time.\\
Step 4: Substituting the given values in the formula, we get distance = 20 km/hr * 2.5 hr = 50 km.\\
So, the person traveled a distance of 50 km.\\
Answer: E\\
\\
\textbf{[Question]}: How many keystrokes are needed to type the numbers from 1 to 500?\\
\textbf{[Options]}: A)1156, B)1392, C)1480, D)1562, E)1788\\
\textbf{[Solution]}: \\
Let's think step by step.\\
Step 1: For 1 to 9, we have 9 numbers, each of which is a single digit, so we need 9 keystrokes.\\
Step 2: For 10 to 99, we have 90 numbers, each of which is a double digit, so we need 90 * 2 = 180 keystrokes.\\
Step 3: For 100 to 499, we have 400 numbers, each of which is a triple digit, so we need 400 * 3 = 1200 keystrokes.\\
Step 4: For 500, this is a triple digit number, so we need 3 keystrokes.\\
Step 5: Adding all these up, we get 9 + 180 + 1200 + 3 = 1392 keystrokes.\\
So, to type the numbers from 1 to 500, we need 1392 keystrokes.\\
Answer: B
\end{tcolorbox}

\subsection{Complex CoT w/ \alignedcot exemplars for AQUA}
\begin{tcolorbox}[breakable, title = {AlignedCoT For GPT-3.5-Turbo}]
\small
\textbf{[Question]}: There were 35 students in a hostel. Due to the admission of 7 new students the expenses of the mess were increased by Rs.84 per day while the average expenditure per head diminished by Re 1. What was the original expenditure of the mess?\\
\textbf{[Options]}: A)Rs 450, B)Rs 920, C)Rs 550, D)Rs.630, E)None of these\\
\textbf{[Solution]}: \\
Let's think step by step.\\
Let's assume the original average expenditure per head be Rs.x\\
So, the total expenditure of the mess before the admission of new students = 35x.\\
After the admission of 7 new students, the total number of students becomes 35 + 7 = 42.\\
The new average expenditure per head = x - 1.\\
So, the total expenditure of the mess after the admission of new students = 42(x - 1).\\
According to the given information, the new total expenditure is increased by Rs.84 per day.\\
So, we can write the equation as: \\
42(x - 1) - 35x = 84.\\
Now, let's solve the equation.\\
42x - 42 - 35x = 84.\\
7x = 126.\\
x = 18.\\
The original expenditure of the mess was Rs. 18 per head.\\
Therefore, the original expenditure of the mess = 35 * 18 = Rs. 630.\\
Answer: D\\
\\
\textbf{[Question]}: A train 200 m long passes a man, running at 5 km/hr in the same direction in which the train is going, in 10 seconds. The speed of the train is?\\
\textbf{[Options]}: A)28, B)50, C)77, D)22, E)12\\
\textbf{[Solution]}: \\
Let's think step by step.\\
Length of the train = 200 m = 0.2 km\\
Speed of the man = 5 km/hr\\
Time taken to pass the man = 10 s = (10 / 3600) hr = (1 / 360) hr\\
Let's assume the speed of the train is 'x' km/hr.\\
Relative speed of the train with respect to the man = (x - 5) km/hr\\
Distance covered by the train in 10 seconds = Length of the train = 200 m\\
Using the formula: Distance = Speed × Time, we can write the equation as: \\
0.2 = (x - 5) × (1 / 360)\\
Now, let's solve the equation.\\
0.2 × 360 = x - 5\\
72 + 5 = x\\
x = 77\\
Therefore, the speed of the train is 77 km/hr.\\
Answer: C\\
\\
\textbf{[Question]}: Solution X contains 20\% of material A and 80\% of material B. Solution Y contains 30\% of material A and 70\% of material B. A mixture of both these solutions contains 22\% of material A in the final product. how much solution X is present in the mixture?\\
\textbf{[Options]}: A)40\%, B)60\%, C)80\%, D)100\%, E)110\%\\
\textbf{[Solution]}: \\
Let's think step by step.\\
Let's assume that the mixture contains x\% of solution X and (100-x)\% of solution Y.\\
In solution X, 20\% of material A is present, so the amount of material A in solution X is 0.2x.\\
In solution Y, 30\% of material A is present, so the amount of material A in solution Y is 0.3(100-x).\\
In the final mixture, 22\% of material A is present, so the amount of material A in the mixture is 0.22(100).\\
Since the amount of material A in the mixture is the sum of the amounts of material A in solution X and solution Y, we can write the equation:\\
0.2x + 0.3(100-x) = 0.22(100)\\
Now, let's solve the equation.\\
0.2x + 30 - 0.3x = 22\\
-0.1x = -8\\
x = 80\\
Therefore, 80\% of solution X is present in the mixture.\\
Answer: C\\
\\
\textbf{[Question]}: A trader sells 40 metres of cloth for Rs.8200 at a profit of Rs.35 per metre of cloth. How much profit will the trder earn on 40 metres of cloth?\\
\textbf{[Options]}: A)Rs.950, B)Rs.1500, C)Rs.1000, D)Rs.1400, E)None of these\\
\textbf{[Solution]}: \\
Let's think step by step.\\
The trader earns a profit of Rs.35 per metre of cloth.\\
The trader sells 40 metres of cloth.\\
So, the trader will earn a profit of Rs.35 x 40 = Rs.1400 on 40 metres of cloth.\\
Answer: D\\
\\
\textbf{[Question]}: If x < y < z and y - x > 5, where x is an even integer and y and z are odd integers, what is the least possible value S of z - x?\\
\textbf{[Options]}: A)6, B)7, C)8, D)9, E)10\\
\textbf{[Solution]}: \\
Let's think step by step.\\
Since x is an even integer and y is an odd integer, the difference y - x will always be an odd integer. \\
Given that y - x > 5, since it has to be an odd integer greater than 5, the smallest possible value for y - x is 7.\\
Since y and z are both odd integers, the difference z - y will always be an even integer.\\
Gicen that y < z, which means z - y > 0. Since z - y is an even integer greater than 0, the smallest possible value for z - y is 2.\\
Therefore, the least possible value of z - x = (z - y) + (y - x) is 2 + 7 = 9.\\
Answer: D\\
\\
\textbf{[Question]}: What is the difference between the C.I. on Rs. 6000 for 1 1/2 years at 4\% per annum compounded yearly and half-yearly?\\
\textbf{[Options]}: A)s.2.04, B)s.2.08, C)s.2.02, D)s.2.83, E)s.2.45\\
\textbf{[Solution]}: \\
Let's think step by step.\\
Let's first give the formula for compound interest:\\
CI = P(1 + r/100)\^{}n - P\\
Where,\\
P = Principal amount\\
r = Rate of interest\\
n = The number of interest periods, which should be an integer\\
Let's calculate the compound interest (CI) for 1 1/2 years at 4\% per annum compounded yearly.\\
We are given that: The principal amount is Rs. 6000. So, P = 6000; The rate of interest is 4\% per annum. So, r = 4\%; The number of interest periods is 1 1/2 years. So n = 1 1/2.\\
The n is not an integer, and people can only get half interest for the rest half year. \\
So, the CI compounded yearly for 1 1/2 years is:\\
CI = P(1 + r/100)(1 + 1/2 * r/100) - P\\
= 6000(1 + 4/100)(1 + 1/2 * 4/100) - 6000\\
= 6000(104/100)(102/100) - 6000\\
= 6000 * 26/25 * 51/50 - 6000\\
= 6000 * 1326/1250 - 6000\\
= 6000 * 1.0608 - 6000\\
= 6364.8 - 6000\\
= 364.8\\
Now, let's calculate the compound interest (CI) for 1 1/2 years at 4\% per annum compounded half-yearly.\\
We are given that: The principal amount is Rs. 6000. So, P = 6000. The rate of interest is 4\% per annum compounded half-yearly. So, r = 4\% / 2 = 2\%. The number of interest periods is (1 1/2) years / (1/2) year = 3.\\
So, the CI compounded half-yearly for 1 1/2 years is:\\
CI = P(1 + r/100)\^{}n - P\\
= 6000(1 + 2/100)\^{}3 - 6000\\
= 6000(102/100)\^{}3 - 6000\\
= 6000 * 51/50 * 51/50 * 51/50 - 6000\\
= 6000 * 132651 / 125000 - 6000\\
= 6000 * 1.061208 - 6000\\
= 6367.248 - 6000\\
= 367.248\\
So, the difference between the two compound interests is: \\
367.248 - 364.8 = 2.448 = Rs.2.45\\
Answer: E\\
\\
\textbf{[Question]}: The average weight of A, B and C is 45 kg. If the average weight of A and B be 40 kg and that of B and C be 45 kg, then the weight of B is:\\
\textbf{[Options]}: A)31 kg, B)32 kg, C)33 kg, D)35 kg, E)None of these\\
\textbf{[Solution]}: \\
Let's think step by step.\\
Let's think step by step.\\
Let the weight of A be x kg, the weight of B be y kg, and the weight of C be z kg.\\
According to the given information, the average weight of A and B is 40 kg. So, we can write the equation:\\
(x + y)/2 = 40\\
Simplifying this equation, we get:\\
x + y = 80\\
Similarly, the average weight of B and C is 45 kg. So, we can write the equation:\\
(y + z)/2 = 45\\
Simplifying this equation, we get:\\
y + z = 90\\
Now, we need to find the weight of B. We can do this by subtracting the weight of A and C from the total weight of A, B, and C.\\
The average weight of A, B, and C is 45 kg. So, we can write the equation:\\
(x + y + z)/3 = 45\\
Simplifying this equation, we get:\\
x + y + z = 135\\
Now, subtracting the equation (x + y = 80) from the equation (x + y + z = 135), we get:\\
z = 55tuting the value of z in the equation (y + z = 90), we get:\\
y + 55 = 90\\
Subtracting 55 from both sides of the equation, we get:\\
y = 35\\
Therefore, the weight of B is 35 kg.\\
Answer: D\\
\\
\textbf{[Question]}: The compound and the simple interests on a certain sum at the same rate of interest for two years are Rs.11730 and Rs.10200 respectively. Find the sum\\
\textbf{[Options]}: A)Rs.17037, B)Rs.17000, C)Rs.17276, D)Rs.170287, E)Rs.171881\\
\textbf{[Solution]}: \\
Let's think step by step.\\
Let's first give the formulas for compound interest (CI) and simple interest (SI).\\
CI = P(1 + r/100)\^{}n - P\\
SI = P * r * n/100\\
Where,\\
P = Principal amount\\
r = Rate of interest\\
n = The number of interest periods, which should be an integer.\\
Given that the compound interest (CI) and the simple interest (SI) for 2 years is Rs. 11730 and Rs. 10200 respectively.\\
So, we have CI = Rs. 11730, SI = Rs. 10200, and n = 2.\\
Substituting these values in the formulas, we have:\\
11730 = P(1 + r/100)\^{}2 - P\\
10200 = P * r * 2/100\\
Now, let's solve the equations to find P.\\
From the second equation, we can rewrite it as:\\
r = 10200 * 100 / (P * 2)\\
r = 510000 / P\\
Now, substitute this value of r in the first equation:\\
11730 = P(1 + 510000/P/100)\^{}2 - P\\
Simplifying the equation, we get:\\
11730 = P(1 + 5100/P)\^{}2 - P\\
11730 = P(1 + 2 * 5100/P + 5100\^{}2/P\^{}2) - P\\
11730 = P + 2* 5100 + 5100\^{}2/P - P\\
11730 = 10200 + 5100\^{}2/P\\
1530 = 5100\^{}2/P\\
P = 5100\^{}2/1530\\
P = 2601000/1530\\
P = 17000\\
Therefore, the sum is Rs. 17000.\\
Answer: B
\end{tcolorbox}

\begin{tcolorbox}[breakable, title = {AlignedCoT For GPT-4}]
\small
\textbf{[Question]}: there were 35 students in a hostel . due to the admission of 7 new students the expenses of the mess were increased by rs . 84 per day while the average expenditure per head diminished by re 1 . what was the original expenditure of the mess ?\\
\textbf{[Options]}: (a) rs 450 (b) rs 920 (c) rs 550 (d) rs . 630 (e) none of these\\
\textbf{[Solution]}: \\
Let's think step by step.\\
The original number of students was 35 and the original expenditure was X. So, the original average expenditure per head was X/35.\\
After the admission of 7 new students, the total number of students became 42 and the total expenditure became X+84. So, the new average expenditure per head was (X+84)/42.\\
According to the problem, the new average expenditure per head is 1 less than the original average expenditure per head. So, we can write the equation as follows: X/35 - (X+84)/42 = 1.\\
Now, let's solving the above equation, \\
Multiplying both sides by 35 * 42, we get 42X - 35(X+84) = 35*42.\\
Simplifying, we get 7X - 35*84 = 35*42.\\
7X = 35*84 + 35*42 = 35(84 + 42) = 35*126 = 35*7*18.\\
X = 35*18 = 630.\\
So, the original expenditure of the mess was Rs. 630.\\
Answer: D\\
\\
\textbf{[Question]}: a train 200 m long passes a man , running at 5 km / hr in the same direction in which the train is going , in 10 seconds . the speed of the train is ?\\
\textbf{[Options]}: (a) 28 (b) 50 (c) 77 (d) 22 (e) 12\\
\textbf{[Solution]}: \\
Let's think step by step.\\
Convert the speed of the man from km/hr to m/s. \\
We know that 1 km/hr = 5/18 m/s. So, the speed of the man is 5*(5/18) = 25/18 m/s.\\
The train passes the man in 10 seconds. This means the relative speed of the train to the man is the distance divided by the time, which is 200m/10s = 20 m/s.\\
The actual speed of the train is the relative speed plus the speed of the man, because they are moving in the same direction. So, the speed of the train is 20 m/s + 25/18 m/s = 385/18 m/s.\\
Convert the speed of the train from m/s to km/hr. \\
We know that 1 m/s = 18/5 km/hr. So, the speed of the train is (385/18)*(18/5) = 77 km/hr.\\
So, the correct answer is 77.\\
Answer: C\\
\\
\textbf{[Question]}: solution x contains 20 \% of material a and 80 \% of material b . solution y contains 30 \% of material a and 70 \% of material b . a mixture of both these solutions contains 22 \% of material a in the final product . how much solution x is present in the mixture ?\\
\textbf{[Options]}: (a) 40 \% (b) 60 \% (c) 80 \% (d) 100 \% (e) 110 \%\\
\textbf{[Solution]}: \\
Let's think step by step.\\
Let's assume that the amount of solution X used is x\% and the amount of solution Y used is (100-x)\%.\\
The amount of material A in the final mixture is given by the sum of the amounts of material A in solutions X and Y. So, we can write the equation as follows: 0.20x + 0.30(100-x) = 22.\\
Now, let's solving the above equation, \\
0.20x + 0.30(100-x) = 22\\
0.20x + 30 - 0.30x = 22\\
0.10x = 8\\
x = 80\\
So, 80\% of solution X is present in the mixture.\\
Answer: C\\
\\
\textbf{[Question]}: a trader sells 40 metres of cloth for rs . 8200 at a profit of rs . 35 per metre of cloth . how much profit will the trder earn on 40 metres of cloth ?\\
\textbf{[Options]}: (a) rs . 950 (b) rs . 1500 (c) rs . 1000 (d) rs . 1400 (e) none of these\\
\textbf{[Solution]}: \\
Let's think step by step.\\
The trader earns a profit of Rs. 35 per metre of cloth.\\
So, for 40 metres of cloth, the trader will earn a profit of 40*35 = Rs. 1400.\\
So, the trader will earn a profit of Rs. 1400 on 40 metres of cloth.\\
Answer: D\\
\\
\textbf{[Question]}: if x < y < z and y - x > 5 , where x is an even integer and y and z are odd integers , what is the least possible value s of z - x ?\\
\textbf{[Options]}: (a) 6 (b) 7 (c) 8 (d) 9 (e) 10\\
\textbf{[Solution]}: \\
Let's think step by step.\\
Since x is an even integer and y is an odd integer, the difference between y and x must be an odd number. Given that y - x > 5, the smallest possible value for y - x is 7.\\
Since y < z and y and z are both odd integers, the smallest possible difference between y and z is 2 (because odd numbers are 2 units apart).\\
Therefore, the smallest possible value for z - x is the sum of the smallest possible values for y - x and y - z, which is 7 + 2 = 9.\\
So, the least possible value of z - x is 9.\\
Answer: D\\
\\
\textbf{[Question]}: what is the difference between the c . i . on rs . 6000 for 1 1 / 2 years at 4 \% per annum compounded yearly and half - yearly ?\\
\textbf{[Options]}: (a) s . 2.04 (b) s . 2.08 (c) s . 2.02 (d) s . 2.83 (e) s . 2.45\\
\textbf{[Solution]}: \\
Let's think step by step.\\
Calculate the compound interest (C.I.) when compounded yearly. \\
The formula for C.I. is A = P(1 + r/n)\^{}(nt), where \\
A is the amount, \\
P is the principal, \\
r is the rate of interest, \\
n is the number of times interest is compounded per time period, \\
and t is the time. \\
Here, P = Rs. 6000, r = 4/100 = 0.04, n = 1 (since it's compounded yearly), and t = 1.5 years. \\
However, the rest half year in the 1.5 years time period can not cover the whole compound period. People can only get half interest for the rest half year.\\
So, A = 6000(1 + 0.04/1)(1 + 0.5*0.04/1)\\
= 6000(1.04)(1.02)\\
= 6000 * 1.0608\\
= 6364.8\\
The C.I. is A - P = Rs. 6364.8 - Rs. 6000 = Rs. 364.8.\\
Calculate the C.I. when compounded half-yearly. Now, n = 2 (since it's compounded half-yearly). \\
So, A = 6000(1 + 0.04/2)\^{}(2*1.5)\\
= 6000(1.02)\^{}3\\
= 6000 * 1.061208\\
= 6367.248\\
The C.I. is A - P = Rs. 6367.248 - Rs. 6000 = Rs. 367.248.\\
The difference between the C.I. when compounded yearly and half-yearly is Rs. 367.248 - Rs. 364.8 = Rs. 2.448. = Rs. 2.45\\
So, the correct answer is Rs. 2.45.\\
Answer: E\\
\\
\textbf{[Question]}: the average weight of a , b and c is 45 kg . if the average weight of a and b be 40 kg and that of b and c be 45 kg , then the weight of b is :\\
\textbf{[Options]}: (a) 31 kg (b) 32 kg (c) 33 kg (d) 35 kg (e) none of these\\
\textbf{[Solution]}: \\
Let's think step by step.\\
The average weight of a, b, and c is 45 kg. So, (a + b + c)/3 = 45 kg.\\
Then, a + b + c = 45*3 = 135 kg.\\
The average weight of a and b is 40 kg. So, (a + b)/2 = 40 kg.\\
Then, a + b = 40*2 = 80 kg.\\
The average weight of b and c is 45 kg. So, (b + c)/2 = 45 kg.\\
Then, b + c = 45*2 = 90 kg.\\
Now, we have a + b = 80 kg and b + c = 90 kg. So, a + b + b + c = 80 + 90 = 170 kg.\\
We already know that a + b + c = 135 kg. So, 135 = 170 - b.\\
Then, b = 170 - 135 = 35 kg.\\
So, the weight of b is 35 kg. \\
Answer: D\\
\\
\textbf{[Question]}: the compound and the simple interests on a certain sum at the same rate of interest for two years are rs . 11730 and rs . 10200 respectively . find the sum\\
\textbf{[Options]}: (a) rs . 17037 (b) rs . 17000 (c) rs . 17276 (d) rs . 170287 (e) rs . 171881\\
\textbf{[Solution]}: \\
Let's think step by step.\\
The difference between the compound interest and the simple interest for two years gives the interest for one year on the simple interest for one year. So, the interest for one year is Rs. 11730 - Rs. 10200 = Rs. 1530.\\
The simple interest for two years is Rs. 10200, so the simple interest for one year is Rs. 10200 / 2 = Rs. 5100.\\
The rate of interest is the interest for one year divided by the principal amount. So, the rate of interest is Rs. 1530 / Rs. 5100 = 0.3 or 30\%.\\
The principal amount is the simple interest for one year divided by the rate of interest. So, the principal amount is Rs. 5100 / 0.3 = Rs. 17000.\\
So, the sum is Rs. 17000.\\
Answer: B
\end{tcolorbox}

\subsection{CoT w/ \alignedcot exemplars for Penguins}
\begin{tcolorbox}[breakable, title = {AlignedCoT For GPT-3.5-Turbo}]
\small
\textbf{Q}: Here is a table where the first line is a header and each subsequent line is a penguin:  name, age, height (cm), weight (kg) Louis, 7, 50, 11 Bernard, 5, 80, 13 Vincent, 9, 60, 11 Gwen, 8, 70, 15  For example: the age of Louis is 7, the weight of Gwen is 15 kg, the height of Bernard is 80 cm.  We now add a penguin to the table:\\
James, 12, 90, 12\\
How many penguins are less than 8 years old?\\
\textbf{Options}:\\
(A) 1\\
(B) 2\\
(C) 3\\
(D) 4\\
(E) 5\\
\textbf{A}: Let's think step by step.\\
We are given a table with the information of several penguins. We need to find out how many penguins are less than 8 years old.\\
Looking at the table, we can see that Louis is 7 years old, Bernard is 5 years old, Vincent is 9 years old, and Gwen is 8 years old. The additional penguin, James, is 12 years old.\\
Out of these five penguins, only Louis and Bernard are less than 8 years old. \\
So the answer is (B).\\
\\
\textbf{Q}: Here is a table where the first line is a header and each subsequent line is a penguin:  name, age, height (cm), weight (kg) Louis, 7, 50, 11 Bernard, 5, 80, 13 Vincent, 9, 60, 11 Gwen, 8, 70, 15  For example: the age of Louis is 7, the weight of Gwen is 15 kg, the height of Bernard is 80 cm.  Which is the youngest penguin?\\
\textbf{Options}:\\
(A) Louis\\
(B) Bernard\\
(C) Vincent\\
(D) Gwen\\
(E) James\\
\textbf{A}: Let's think step by step.\\
We are given a table with the information of several penguins. We need to find out the the youngest penguin.\\
First, let's identify the column that represents the age of each penguin. In this case, it is the second column.\\
Next, let's compare the ages of each penguin:\\
Louis: 7\\
Bernard: 5\\
Vincent: 9\\
Gwen: 8\\
From the comparison, we can see that Bernard is the youngest penguin with an age of 5.\\
So the answer is (B).\\
\\
\textbf{Q}: Here is a table where the first line is a header and each subsequent line is a penguin:  name, age, height (cm), weight (kg) Louis, 7, 50, 11 Bernard, 5, 80, 13 Vincent, 9, 60, 11 Gwen, 8, 70, 15  For example: the age of Louis is 7, the weight of Gwen is 15 kg, the height of Bernard is 80 cm.  What is the name of the second penguin sorted by alphabetic order?\\
\textbf{Options}:\\
(A) Louis\\
(B) Bernard\\
(C) Vincent\\
(D) Gwen\\
(E) James\\
\textbf{A}: Let's think step by step.\\
We are given a table with the information of several penguins. We need to find the name of the second penguin sorted by alphabetic order.\\
First, let's identify the name of each penguin in the table. In this case, the names are in the first column. So, the names are Louis, Bernard, Vincent, and Gwen.\\
Next, let's sort the penguins by alphabetical order. The penguins listed in alphabetical order are Bernard, Gwen, Louis, and Vincent.\\
Therefore, the name of the second penguin sorted by alphabetical order is Gwen.\\
So the answer is (D).
\end{tcolorbox}

\begin{tcolorbox}[breakable, title = {AlignedCoT For GPT-4}]
\small
\textbf{Q}: Here is a table where the first line is a header and each subsequent line is a penguin:  name, age, height (cm), weight (kg) Louis, 7, 50, 11 Bernard, 5, 80, 13 Vincent, 9, 60, 11 Gwen, 8, 70, 15  For example: the age of Louis is 7, the weight of Gwen is 15 kg, the height of Bernard is 80 cm.  We now add a penguin to the table:\\
James, 12, 90, 12\\
How many penguins are less than 8 years old?\\
\textbf{Options}:\\
(A) 1\\
(B) 2\\
(C) 3\\
(D) 4\\
(E) 5\\
\textbf{A}: Let's think step by step.\\
First, we need to find the age of each penguin. \\
According to the table, Louis is 7 years old, Bernard is 5 years old, Vincent is 9 years old, Gwen is 8 years old, and James is 12 years old.\\
We are looking for penguins that are less than 8 years old. \\
Louis is 7 years old, which is less than 8. Bernard is 5 years old, which is also less than 8. \\
Therefore, there are 2 penguins that are less than 8 years old. \\
So the answer is (B).\\
\\
\textbf{Q}: Here is a table where the first line is a header and each subsequent line is a penguin:  name, age, height (cm), weight (kg) Louis, 7, 50, 11 Bernard, 5, 80, 13 Vincent, 9, 60, 11 Gwen, 8, 70, 15  For example: the age of Louis is 7, the weight of Gwen is 15 kg, the height of Bernard is 80 cm.  Which is the youngest penguin?\\
\textbf{Options}:\\
(A) Louis\\
(B) Bernard\\
(C) Vincent\\
(D) Gwen\\
(E) James\\
\textbf{A}: Let's think step by step.\\
First, we need to find the age of each penguin. \\
According to the table, Louis is 7 years old, Bernard is 5 years old, Vincent is 9 years old, and Gwen is 8 years old. \\
The youngest penguin is the one with the smallest age. In this case, the 5 years old Bernard is the youngest penguin. \\
So the answer is (B).\\
\\
\textbf{Q}: Here is a table where the first line is a header and each subsequent line is a penguin:  name, age, height (cm), weight (kg) Louis, 7, 50, 11 Bernard, 5, 80, 13 Vincent, 9, 60, 11 Gwen, 8, 70, 15  For example: the age of Louis is 7, the weight of Gwen is 15 kg, the height of Bernard is 80 cm.  What is the name of the second penguin sorted by alphabetic order?\\
\textbf{Options}:\\
(A) Louis\\
(B) Bernard\\
(C) Vincent\\
(D) Gwen\\
(E) James\\
\textbf{A}: Let's think step by step.\\
First, we need to find the age of each penguin. \\
According to the table, the names of the penguins are Louis, Bernard, Vincent, and Gwen. \\
Then, we need to sort the names of the penguins in alphabetical order. we get Bernard, Gwen, Louis, and Vincent. \\
The second penguin in this sorted list is Gwen. \\
So the answer is (D).
\end{tcolorbox}

\subsection{Complex CoT w/ \alignedcot exemplars for Penguins}
\begin{tcolorbox}[breakable, title = {AlignedCoT For GPT-3.5-Turbo}]
\small
\textbf{Q}: Here is a table where the first line is a header and each subsequent line is a penguin:  name, age, height (cm), weight (kg) Louis, 7, 50, 11 Bernard, 5, 80, 13 Vincent, 9, 60, 11 Gwen, 8, 70, 15  For example: the age of Louis is 7, the weight of Gwen is 15 kg, the height of Bernard is 80 cm.  We now add a penguin to the table:\\
James, 12, 90, 12\\
And here is a similar table, but listing giraffes:\\
name, age, height (cm), weight (kg)\\
Jody, 5, 430, 620\\
Gladys, 10, 420, 590\\
Marian, 2, 310, 410\\
Donna, 9, 440, 650\\
What is the name of the last animal sorted by alphabetic order?\\
\textbf{Options}:\\
(A) Vincent\\
(B) Louis\\
(C) Louise\\
(D) Marian\\
(E) Jody\\
\textbf{A}: Let's think step by step.\\
First, let's identify the name of each animal in the tables. In this case, the names are in the first column. \\
For the penguins table, the names are Louis, Bernard, Vincent, Gwen, and James.\\
For the giraffes table, the names are Jody, Gladys, Marian, and Donna.\\
Next, let's sort the names in alphabetical order.\\
For the penguins table, the names in alphabetical order are Bernard, Gwen, James, Louis, and Vincent.\\
For the giraffes table, the names in alphabetical order are Donna, Gladys, Jody, and Marian.\\
Now, let's find the last animal in alphabetical order. \\
For the penguins table, the last animal in alphabetical order is Vincent.\\
For the giraffes table, the last animal in alphabetical order is Marian.\\
Vincent is alphabetically after Marian.\\
So the answer is (A).\\
\\
\textbf{Q}: Here is a table where the first line is a header and each subsequent line is a penguin:  name, age, height (cm), weight (kg) Louis, 7, 50, 11 Bernard, 5, 80, 13 Vincent, 9, 60, 11 Gwen, 8, 70, 15  For example: the age of Louis is 7, the weight of Gwen is 15 kg, the height of Bernard is 80 cm.  We now add a penguin to the table:\\
James, 12, 90, 12\\
And here is a similar table, but listing giraffes:\\
name, age, height (cm), weight (kg)\\
Jody, 5, 430, 620\\
Gladys, 10, 420, 590\\
Marian, 2, 310, 410\\
Donna, 9, 440, 650\\
How many animals are less than 8 years old? \\
\textbf{Options}:\\
(A) 2\\
(B) 3\\
(C) 4\\
(D) 5\\
(E) 6\\
\textbf{A}: Let's think step by step.\\
First, let's count the number of penguins that are less than 8 years old. \\
From the given table, we can see that Louis is 7 years old, Bernard is 5 years old, Vincent is 9 years old, and Gwen is 8 years old. \\
Out of these four penguins, only Louis and Bernard are less than 8 years old. \\
So, there are 2 penguins that are less than 8 years old. \\
Next, let's count the number of giraffes that are less than 8 years old. \\
From the given table, we can see that Jody is 5 years old, Gladys is 10 years old, Marian is 2 years old, and Donna is 9 years old. \\
Out of these four giraffes, only Jody and Marian are less than 8 years old. \\
So, there are 2 giraffes that are less than 8 years old. \\
Adding the number of penguins and giraffes that are less than 8 years old, we get: \\
2 (penguins) + 2 (giraffes) = 4\\
Therefore, there are 4 animals in total that are less than 8 years old. \\
So the answer is (C).\\
\\
\textbf{Q}: Here is a table where the first line is a header and each subsequent line is a penguin:  name, age, height (cm), weight (kg) Louis, 7, 50, 11 Bernard, 5, 80, 13 Vincent, 9, 60, 11 Gwen, 8, 70, 15  For example: the age of Louis is 7, the weight of Gwen is 15 kg, the height of Bernard is 80 cm.  \\
We now add two penguins to the table:\\
James, 4, 90, 12, Dick, 5, 75, 15.\\
Which is the youngest penguin?\\
\textbf{Options}:\\
(A) Louis\\
(B) Bernard\\
(C) Vincent\\
(D) Gwen\\
(E) James\\
\textbf{A}: Let's think step by step.\\
First, let's identify the age of each penguin.\\
Louis is 7 years old, Bernard is 5 years old, Vincent is 9 years old, Gwen is 8 years old, James is 4 years old, and Dick is 5 years old. \\
Out of these options, James is the youngest penguin as he is 4 years old. \\
Therefore, the answer is (E) James.
\end{tcolorbox}

\begin{tcolorbox}[breakable, title = {AlignedCoT For GPT-4}]
\small
\textbf{Q}: Here is a table where the first line is a header and each subsequent line is a penguin:  name, age, height (cm), weight (kg) Louis, 7, 50, 11 Bernard, 5, 80, 13 Vincent, 9, 60, 11 Gwen, 8, 70, 15  For example: the age of Louis is 7, the weight of Gwen is 15 kg, the height of Bernard is 80 cm.  We now add a penguin to the table:\\
James, 12, 90, 12\\
And here is a similar table, but listing giraffes:\\
name, age, height (cm), weight (kg)\\
Jody, 5, 430, 620\\
Gladys, 10, 420, 590\\
Marian, 2, 310, 410\\
Donna, 9, 440, 650\\
What is the name of the last animal sorted by alphabetic order?\\
\textbf{Options}:\\
(A) Vincent\\
(B) Louis\\
(C) Louise\\
(D) Marian\\
(E) Jody\\
\textbf{A}: Let's think step by step.\\
First, let's list all the names of the animals: Louis, Bernard, Vincent, Gwen, James, Jody, Gladys, Marian, Donna.\\
When we sort these names in alphabetical order, we get: Bernard, Donna, Gladys, Gwen, James, Jody, Louis, Marian, Vincent. \\
So, the last animal sorted by alphabetic order is Vincent. \\
Therefore, the correct answer is (A) Vincent.\\
\\
\textbf{Q}: Here is a table where the first line is a header and each subsequent line is a penguin:  name, age, height (cm), weight (kg) Louis, 7, 50, 11 Bernard, 5, 80, 13 Vincent, 9, 60, 11 Gwen, 8, 70, 15  For example: the age of Louis is 7, the weight of Gwen is 15 kg, the height of Bernard is 80 cm.  We now add a penguin to the table:\\
James, 12, 90, 12\\
And here is a similar table, but listing giraffes:\\
name, age, height (cm), weight (kg)\\
Jody, 5, 430, 620\\
Gladys, 10, 420, 590\\
Marian, 2, 310, 410\\
Donna, 9, 440, 650\\
How many animals are less than 8 years old? \\
\textbf{Options}:\\
(A) 2\\
(B) 3\\
(C) 4\\
(D) 5\\
(E) 6\\
\textbf{A}: Let's think step by step.\\
First, let's look at the penguins. Louis is 7 years old, Bernard is 5 years old, Vincent is 9 years old, Gwen is 8 years old, and James is 12 years old. So, there are 2 penguins that are less than 8 years old: Louis and Bernard.\\
Now, let's look at the giraffes. Jody is 5 years old, Gladys is 10 years old, Marian is 2 years old, and Donna is 9 years old. So, there are 2 giraffes that are less than 8 years old: Jody and Marian.\\
Adding these together, there are 2 penguins + 2 giraffes = 4 animals that are less than 8 years old.\\
So, the answer is (C) 4.\\
\\
\textbf{Q}: Here is a table where the first line is a header and each subsequent line is a penguin:  name, age, height (cm), weight (kg) Louis, 7, 50, 11 Bernard, 5, 80, 13 Vincent, 9, 60, 11 Gwen, 8, 70, 15  For example: the age of Louis is 7, the weight of Gwen is 15 kg, the height of Bernard is 80 cm.  \\
We now add two penguins to the table:\\
James, 4, 90, 12, Dick, 5, 75, 15.\\
Which is the youngest penguin?\\
\textbf{Options}:\\
(A) Louis\\
(B) Bernard\\
(C) Vincent\\
(D) Gwen\\
(E) James\\
\textbf{A}: Let's think step by step.\\
First, we need to look at the age of each penguin. Louis is 7, Bernard is 5, Vincent is 9, Gwen is 8, James is 4, and Dick is 5. The youngest penguin is James, who is 4 years old. \\
So, the answer is (E) James.
\end{tcolorbox}

\subsection{Style Conversion Prompt for GSM8K}
In our experiment, we use GPT-3.5-Turbo to convert the training set in GSM8K.
During data preprocessing, we deleted the computational part (bracketed by '<<' and '>>') of the training set.
\begin{tcolorbox}[breakable, title = {Style Conversion Prompt}]
\small
\textbf{[Question]}: Angelo and Melanie want to plan how many hours over the next week they should study together for their test next week. They have 2 chapters of their textbook to study and 4 worksheets to memorize. They figure out that they should dedicate 3 hours to each chapter of their textbook and 1.5 hours for each worksheet. If they plan to study no more than 4 hours each day, how many days should they plan to study total over the next week if they take a 10-minute break every hour, include 3 10-minute snack breaks each day, and 30 minutes for lunch each day?\\
\textbf{[Original Solution]}:\\
Let's think step by step.\\
Angelo and Melanie think they should dedicate 3 hours to each of the 2 chapters, 3 hours x 2 chapters = 6 hours total.\\
For the worksheets they plan to dedicate 1.5 hours for each worksheet, 1.5 hours x 4 worksheets = 6 hours total.\\
However, they need to include time for breaks and lunch. Every hour they want to include a 10-minute break, so 4 total hours x 10 minutes = 40 extra minutes for breaks.\\
They also want to include 3 10-minute snack breaks everyday, 3 x 10 minutes = 30 minutes.\\
And they want to include 30 minutes for lunch each day, so 40 minutes for breaks + 30 minutes for snack breaks + 30 minutes for lunch = 100 minutes, or 100 / 60 minutes per hour = 1.67 extra hours.\\
So, the total time they can study each day is 4 - 1.67 = 2.33 hours.\\
So they need 12 hours / 2.33 hours each day = 5.15 days to study.\\
Rounding up, the result is 6, so they need 6 days to study.\\
Answer: 6\\
\textbf{[Converted Solution]}:\\
Let's think step by step.\\
STEP 1: Calculate the total time needed to study the textbook chapters.\\
They have 2 chapters of their textbook to study, and they plan to dedicate 3 hours to each chapter.\\
So, the total time needed to study the textbook chapters is 2 * 3 = 6 hours.\\
STEP 2: Calculate the total time needed to memorize the worksheets.\\
They have 4 worksheets to memorize, and they plan to dedicate 1.5 hours for each worksheet.\\
So, the total time needed to memorize the worksheets is 4 * 1.5 = 6 hours.\\
STEP 3: Calculate the total time needed for breaks and meals each day.\\
They take a 10-minute break every hour, so in a 4-hour study session, they will have 4 * 10 = 40 minutes of breaks.\\
They also have 3 10-minute snack breaks each day, so they will have 3 * 10 = 30 minutes of snack breaks.\\
They have 30 minutes for lunch each day.\\
So, the total time needed for breaks and meals each day is 40 + 30 + 30 = 100 minutes, which is equal to 100/60 = 1.67 hours.\\
STEP 4: Calculate the total time needed to study each day.\\
They plan to study no more than 4 hours each day, and they need to subtract the time needed for breaks and meals.\\
So, the total time they can study each day is 4 - 1.67 = 2.33 hours.\\
STEP 5: Calculate the total number of days needed to study.\\
They need a total of 6 hours to study the textbook chapters and 6 hours to memorize the worksheets.\\
So, the total time needed to study is 6 + 6 = 12 hours.\\
Since they can study 2.33 hours each day, they will need 12 / 2.33 = 5.15 days to complete their study.\\
Since they can't study for a fraction of a day, they should plan to study for 6 days.\\
Therefore, they should plan to study for a total of 6 days over the next week.\\
Answer: 6\\
\\
\textbf{[Question]}: Mark's basketball team scores 25 2 pointers, 8 3 pointers and 10 free throws.  Their opponents score double the 2 pointers but half the 3 pointers and free throws.  What's the total number of points scored by both teams added together?\\
\textbf{[Original Solution]}:\\
Let's think step by step.\\
Mark's team scores 25 2 pointers, meaning they scored 25*2= 50 points in 2 pointers.\\
His team also scores 6 3 pointers, meaning they scored 8*3= 24 points in 3 pointers\\
They scored 10 free throws, and free throws count as one point so they scored 10*1=10 points in free throws.\\
All together his team scored 50+24+10= 84 points\\
Mark's opponents scored double his team's number of 2 pointers, meaning they scored 50*2=100 points in 2 pointers.\\
His opponents scored half his team's number of 3 pointers, meaning they scored 24/2= 12 points in 3 pointers.\\
They also scored half Mark's team's points in free throws, meaning they scored 10/2=5 points in free throws.\\
All together Mark's opponents scored 100+12+5=117 points\\
The total score for the game is both team's scores added together, so it is 84+117=201 points\\
Answer: 201\\
\textbf{[Converted Solution]}:\\
Let's think step by step.\\
STEP 1: Calculate the number of points scored by Mark's team.\\
Mark's team scores 25 2 pointers, which means they score 25 * 2 = 50 points from 2 pointers.\\
They also score 8 3 pointers, which means they score 8 * 3 = 24 points from 3 pointers.\\
Finally, they score 10 free throws, which means they score 10 * 1 = 10 points from free throws.\\
Therefore, the total number of points scored by Mark's team is 50 + 24 + 10 = 84 points.\\
STEP 2: Calculate the number of points scored by the opponents.\\
The opponents score double the 2 pointers, which means they score 2 * 50 = 100 points from 2 pointers.\\
They also score half the 3 pointers, which means they score 0.5 * 24 = 12 points from 3 pointers.\\
Finally, they score half the free throws, which means they score 0.5 * 10 = 5 points from free throws.\\
Therefore, the total number of points scored by the opponents is 100 + 12 + 5 = 117 points.\\
STEP 3: Calculate the total number of points scored by both teams.\\
The total number of points scored by both teams is 84 + 117 = 201 points.\\
Answer: 201\\
\\
\textbf{[Question]}: A group of 4 fruit baskets contains 9 apples, 15 oranges, and 14 bananas in the first three baskets and 2 less of each fruit in the fourth basket. How many fruits are there?\\
\textbf{[Original Solution]}:\\
Let's think step by step.\\
For the first three baskets, the number of apples and oranges in one basket is 9+15=24\\
In total, together with bananas, the number of fruits in one basket is 24+14=38 for the first three baskets.\\
Since there are three baskets each having 38 fruits, there are 3*38=114 fruits in the first three baskets.\\
The number of apples in the fourth basket is 9-2=7\\
There are also 15-2=13 oranges in the fourth basket\\
The combined number of oranges and apples in the fourth basket is 13+7=20\\
The fourth basket also contains 14-2=12 bananas.\\
In total, the fourth basket has 20+12=32 fruits.\\
The four baskets together have 32+114=146 fruits.\\
Answer: 146\\
\textbf{[Converted Solution]}:\\
Let's think step by step.\\
STEP 1: Figure out the total number of fruits in the first three baskets.\\
In each of the first three basket, there are 9 apples + 15 oranges + 14 bananas = 38 fruits.\\
So, in the first three baskets, there are a total of 38 fruits * 3 = 114 fruits.\\
STEP 2: Figure out the number of fruits in the fourth basket.\\
In the fourth basket, there are 2 less of each fruit compared to the first three baskets.\\
So, there are 9 apples - 2 = 7 apples.\\
There are 15 oranges - 2 = 13 oranges.\\
There are 14 bananas - 2 = 12 bananas.\\
Therefore, in the fourth basket, there are a total of 7 apples + 13 oranges + 12 bananas = 32 fruits.\\
STEP 3: Figure out the total number of fruits in all four baskets.\\
The four baskets together have 114 (total fruits in the first three baskets) + 32 (fruits in the fourth basket) = 146 fruits.\\
Answer: 146\\
\\
\textbf{[Question]}: In a certain school, 2/3 of the male students like to play basketball, but only 1/5 of the female students like to play basketball. What percent of the population of the school do not like to play basketball if the ratio of the male to female students is 3:2 and there are 1000 students?\\
\textbf{[Original Solution]}:\\
Let's think step by step.\\
The students are divided into 3 + 2 = 5 parts where 3 parts are for males and 2 parts are for females.\\
Each part represents 1000/5 = 200 students.\\
So, there are 3 x 200 = 600 males.\\
And there are 2 x 200 = 400 females.\\
Hence, 600 x 2/3 = 400 males play basketball.\\
And 400 x 1/5 = 80 females play basketball.\\
A total of 400 + 80 = 480 students play basketball.\\
Therefore, 1000 - 480 = 520 do not like to play basketball.\\
The percentage of the school that do not like to play basketball is 520/1000 * 100 = 52\\
Answer: 52\\
\textbf{[Converted Solution]}:\\
Let's think step by step.\\
STEP 1: Figure out the number of male and female students in the school.\\
The students are divided into 3 + 2 = 5 parts where 3 parts are for males and 2 parts are for females.\\
Each part represents 1000/5 = 200 students.\\
So, there are 3 x 200 = 600 males.\\
And there are 2 x 200 = 400 females.\\
STEP 2: Figure out the number of males and females who like to play basketball.\\
2/3 of the male students like to play basketball. Hence, 600 x 2/3 = 400 males play basketball.\\
1/5 of the female students like to play basketball. Hence, 400 x 1/5 = 80 females play basketball.\\
STEP 3: Figure out how many people do not like to play basketball.\\
A total of 400 + 80 = 480 students play basketball.\\
Therefore, 1000 - 480 = 520 do not like to play basketball.\\
STEP 4: Calculate the percentage of the population that do not like to play basketball.\\
The percentage of the school that do not like to play basketball is 520/1000 * 100 = 52\\
Answer: 52
\end{tcolorbox}

\end{document}